\PassOptionsToPackage{prologue,dvipsnames,table}{xcolor}
\documentclass[10pt,twocolumn,letterpaper]{article}

\usepackage[pagenumbers]{cvpr} 
\usepackage{cvprabbrv}
\usepackage{multirow}
\usepackage{graphicx}
\usepackage{xcolor} 

\definecolor{cvprblue}{rgb}{0.21,0.49,0.74}
\usepackage[pagebackref,breaklinks,colorlinks,allcolors=cvprblue]{hyperref}

\definecolor{hotpink}{cmyk}{0,0.78,0.18,0}  
\hypersetup{
  urlcolor=hotpink
}

\def\paperID{3083} 
\def\confName{CVPR}
\def\confYear{2025}

\title{Change3D: Revisiting Change Detection and Captioning from A Video Modeling Perspective}

\author{
Duowang Zhu$^{1}$,~~Xiaohu Huang$^{2}$,~~Haiyan Huang$^1$,~~Hao Zhou$^3$,~~and Zhenfeng Shao$^{1}$\thanks{Corresponding author: shaozhenfeng@whu.edu.cn.}\\
$^1$Wuhan University~~$^2$The University of Hong Kong~~$^3$Bytedance\\
\url{https://zhuduowang.github.io/Change3D}
}

\begin{document}
\maketitle

\begin{abstract}
In this paper, we present Change3D, a framework that reconceptualizes the change detection and captioning tasks through video modeling. Recent methods have achieved remarkable success by regarding each pair of bi-temporal images as separate frames. They employ a shared-weight image encoder to extract spatial features and then use a change extractor to capture differences between the two images. However, image feature encoding, being a task-agnostic process, cannot attend to changed regions effectively. Furthermore, different change extractors designed for various change detection and captioning tasks make it difficult to have a unified framework. To tackle these challenges, Change3D regards the bi-temporal images as comprising two frames akin to a tiny video. By integrating learnable perception frames between the bi-temporal images, a video encoder enables the perception frames to interact with the images directly and perceive their differences. Therefore, we can get rid of the intricate change extractors, providing a unified framework for different change detection and captioning tasks. We verify Change3D on multiple tasks, encompassing change detection (including binary change detection, semantic change detection, and building damage assessment) and change captioning, across eight standard benchmarks. Without bells and whistles, this simple yet effective framework can achieve superior performance with an ultra-light video model comprising only $\sim$6\%-13\% of the parameters and $\sim$8\%-34\% of the FLOPs compared to state-of-the-art methods. We hope that Change3D could be an alternative to 2D-based models and facilitate future research. 
\end{abstract}

\begin{figure}[htb]
\centering
\includegraphics[width=\linewidth]{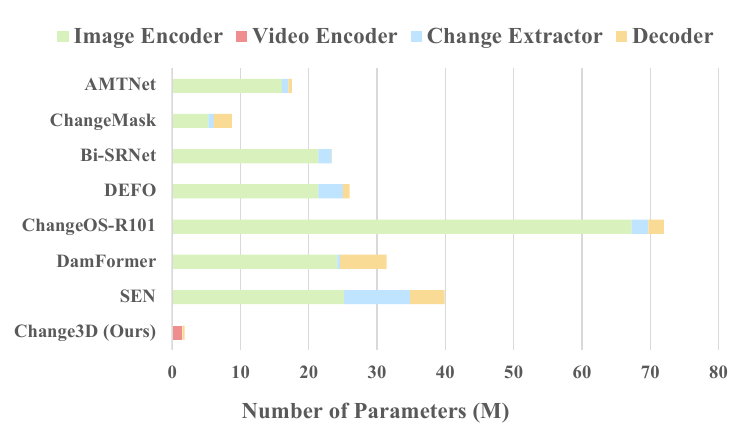}
\caption{The parameter distribution in existing change detection and captioning methods indicates most parameters focused on image encoding, with few allocated to change extraction. This imbalance suggests an insufficient emphasis on task-related parameter learning. In contrast, our approach primarily focuses on video encoding, a task-specific process that effectively extracts changes.
}
\label{fig:parameter_distribution_stacked_chart}
\vspace{-0.8em}
\end{figure}

\begin{figure*}[tb]
\centering
\includegraphics[width=0.9\linewidth]{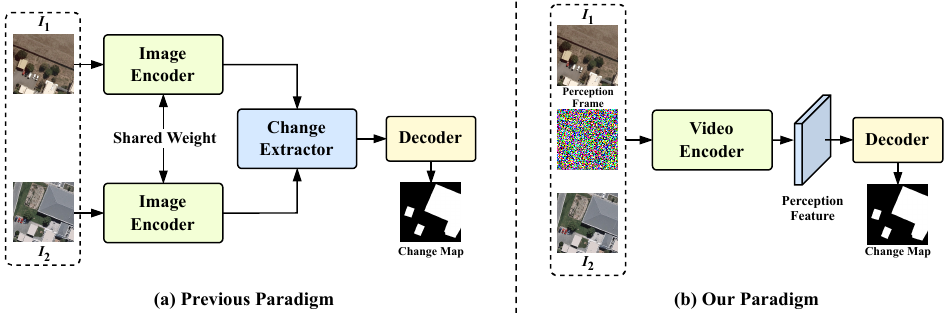}
\caption{\textbf{Previous paradigm \textit{vs.} our paradigm.} (a) Previous paradigm treats bi-temporal image pairs as separate inputs, where each is processed individually by a shared-weight image encoder to extract spatial features, followed by a dedicated change extractor to capture differences and a decoder to make predictions. (b) Our proposed paradigm rethinks the change detection and captioning tasks from a video modeling perspective. By incorporating a learnable perception frame between the bi-temporal images, a video encoder facilitates direct interaction between the perception frame and images to extract differences, eliminating the need for intricate change extractors and providing a unified framework for multiple tasks.
}
\label{fig:different_paradigms}
\end{figure*}

\section{Introduction}
\label{sec:introduction}


Change detection and captioning play a crucial role in the remote sensing field, employing pairs of bi-temporal images taken from the same geographic area at different times to track and analyze changes on the Earth’s surface over time \citep{radke2005image}. It has been widely applied in various applications such as damage assessment \citep{zheng2021building, chen2022dual_DamFormer}, urban planning \citep{wang2021land}, arable land protection \citep{lunetta2022land}, and environmental management \citep{kennedy2009remote}. Existing studies have explored various 2D models for bi-temporal change detection and captioning modeling, such as Convolutional Neural Networks (CNNs)-based \citep{zhang2023global_GASNet, feng2023change_DMINet, wu2023fully_FCD-GN, zheng2022changemask_ChangeMask}, Transformer-based \citep{bandara2022transformer_ChangeFormer, zhang2022swinsunet_SwinSUNet, chen2022dual_DamFormer, liu2023decoupling_PromptCC}, and their hybrid models \citep{chen2021remote_BIT, liu2023attention_AMTNet, zheng2022changemask_ChangeMask}. As shown in Fig.~\ref{fig:different_paradigms} (a), most of the methods typically follow such a paradigm: (1) A pair of bi-temporal images are treated as distinct inputs, processing each image individually through a shared-weight image encoder (\eg, ResNet \citep{he2016deep_ResNet}, UNet \citep{ronneberger2015u_unet}, Swin Transformer \citep{liu2021swin}) for spatial feature extraction. (2) Then, a well-designed change extractor, primarily leveraging attention mechanisms \citep{vaswani2017attention} like cross-attention \citep{chen2021crossvit} or spatial-channel attention \citep{woo2018cbam}, is introduced for bi-temporal feature interaction to detect the differences between them. (3) Finally, a decoder is employed to restore the feature map to its original resolution.


While encouraging results have been observed, these methods have the following issues: (1) The image encoder is designed to learn spatial features independently for bi-temporal images, lacking task-specificity in modeling changes. Although a previous method \citep{fang2023changer_Changer} introduces inter-frame feature interaction during image encoding, it cannot extract changes and still needs a dedicated change extractor. Moreover, as shown in Fig.~\ref{fig:parameter_distribution_stacked_chart}, this paradigm results in an unreasonable parameter distribution, with the majority focusing on independent image encoding and few dedicating to change extraction, without emphasizing the parameter learning for change detection and captioning tasks. (2) Current methods necessitate separate design considerations for change extractors in different tasks, hindering the utilization of a unified framework for addressing all change detection and captioning tasks effectively. Inspired by video modeling approaches \citep{feichtenhofer2019slowfast_SlowFast, feichtenhofer2020x3d_X3D, li2022uniformer_UniFormerV1} that they can effectively model the relations between images, we seek to revisit the bi-temporal images as a tiny video, leveraging the spatiotemporal learning capability inherent in video models to capture changes.

To this end, we propose a unified framework named \textbf{Change3D}, serving as a competitive alternative to 2D-model-based approaches. Our key idea is to rethink the extraction of bi-temporal changes through the lens of video modeling techniques. As depicted in Fig.~\ref{fig:different_paradigms} (b), we ingeniously redefine bi-temporal change detection and captioning as a task akin to video modeling by incorporating a learnable frame that can learn to perceive. In particular, we create a 3D volume by aligning bi-temporal images with the learnable perception frame, all arranged along the time dimension. Change3D then utilizes a video encoder to process this input, enabling the perception frame to observe and discern the differences between the bi-temporal images. Ultimately, the features derived from the perception frame are input into a decoder, which generates a probability map or caption indicating the changes. Change3D can tackle the challenges of bi-temporal image-based methods stated above. First, the distribution of model parameters in Change3D primarily focuses on video feature encoding, a task-specific process adept at extracting changes effectively. Second, intricate change extractors can be eliminated, offering a unified framework for various change detection and captioning tasks.

Extensive experiments on four change detection and captioning tasks (\ie, binary change detection, semantic change detection, building damage assessment, and change captioning) and eight widely recognized datasets (\ie, LEVIR-CD \citep{chen2020spatial_levir_cd}, WHU-CD \citep{ji2018fully_whu_cd}, CLCD \citep{liu2022cnn_clcd}, HRSCD \citep{daudt2019multitask_HRSCD}, SECOND \citep{yang2021asymmetric_ASN_SECOND}, xBD \citep{gupta2019creating_xBD}, LEVIR-CC \citep{liu2022remote_RSICCformer}, and DUBAI-CC \citep{hoxha2022change_DUBAI}) show that Change3D achieves state-of-the-art performance across the board. Meanwhile, we employ various video models including CNN-based (\eg, I3D \citep{carreira2017quo_I3D}, SlowFast \citep{feichtenhofer2019slowfast_SlowFast}, X3D \citep{feichtenhofer2020x3d_X3D}), and Transformer-based (\eg, UniFormer \citep{li2022uniformer_UniFormerV1}) with different pre-trained weights (\eg, AVA \citep{gu2018ava_AVA}, Charades \citep{sigurdsson2016hollywood_Charades}, Something-Something V2 \citep{goyal2017something_something_something_v2}, and Kinetics-400 \citep{kay2017kinetics_KINETICS}) to confirm the efficacy of the proposed paradigm. Notably, despite the simplicity of the proposed method compared to advanced 2D models, Change3D outperforms methods that utilize these complex structural designs, showcasing the effectiveness of the proposed paradigm for bi-temporal change detection and captioning tasks.

The key contributions of this paper are summarized as follows:

\begin{itemize}[leftmargin=2em]
\item We revisit the bi-temporal change detection and captioning tasks from a video modeling perspective and introduce a paradigm called Change3D, a simple yet effective 3D modeling approach for bi-temporal change detection and captioning.

\item By injecting the perception frame into the inputs, Change3D efficiently extracts changes using 3D models without the need for elaborate structural designs or complex frameworks.

\item Change3D achieves state-of-the-art performance with an efficient video model (\ie, X3D-L) across multiple benchmarks, demonstrating the superiority of the proposed method. Moreover, both quantitative and qualitative analyses validate the efficacy of the new paradigm we have introduced, further solidifying the effectiveness of our approach.
\end{itemize}

\section{Related Work}
\label{sec:related_work}
\textbf{Change Detection \& Captioning.}
CNN-based approaches have been the mainstream framework in the literature \citep{zhang2023global_GASNet, feng2022icif_ICIF, fang2023changer_Changer, daudt2018fully_FCEF, feng2023change_DMINet, feng2023lightweight_LCANet, liu2020building_DTCDSCN, daudt2018urban, chang2024triple_JFRNet, li2024decoder_DEFO, ding2022bi_Bi-SRNet, tian2023temporal_TCRPN, liu2022remote_RSICCformer} for a long time, known for their hierarchical feature modeling capabilities. These works primarily focus on multi-scale feature extraction, difference modeling, foreground-background class imbalance, \etc. For instance, methods in \citep{daudt2018fully_FCEF, daudt2018urban, daudt2019multitask_HRSCD} utilize fully convolutional networks to capture hierarchical features for multi-scale feature representations. For adequate differential feature modeling, approaches in \citep{feng2022icif_ICIF, feng2023change_DMINet, chang2024triple_JFRNet, ding2022bi_Bi-SRNet, tian2023temporal_TCRPN, wang2024pcdasnet_PCDASNet, li2024decoder_DEFO, lin2022transition_P2VCD, fang2023changer_Changer} incorporate the attention mechanism \citep{vaswani2017attention} or feature exchange to establish relational dependencies between bi-temporal features. Several studies \citep{zhang2023global_GASNet, zhang2023aernet_AERNet} address the significant challenge posed by foreground-background class imbalance by developing innovative loss functions. Motivated by the achievements of Vision Transformer \citep{dosovitskiy2020image_ViT} and its variants \citep{liu2021swin, liu2022swin, wang2021pyramid_pvt} for various visual tasks \citep{li2022exploring_ViTDet, yao2024vitmatte, xu2022vitpose, chen2023vision_ViTAdapter}, several works \citep{chen2021remote_BIT, bandara2022transformer_ChangeFormer, liu2023attention_AMTNet, ma2024eatder_EATDer, zhang2022swinsunet_SwinSUNet, jiang2023vct_VcT, ding2024joint_SCanNet, zheng2022changemask_ChangeMask, chen2022dual_DamFormer, liu2023decoupling_PromptCC} have explored the application of transformers in change detection and captioning tasks. Some of these methods \citep{zhang2022swinsunet_SwinSUNet, bandara2022transformer_ChangeFormer, chen2022dual_DamFormer, cui2023mtscd_MTSCD, zhu2024changevit_ChangeViT} utilize pure transformers, while others \citep{chen2021remote_BIT, liu2023attention_AMTNet, ma2024eatder_EATDer, jiang2023vct_VcT, ding2024joint_SCanNet, zheng2022changemask_ChangeMask} adopt CNN-Transformer hybrid architectures. Methods in \citep{zhang2022swinsunet_SwinSUNet, bandara2022transformer_ChangeFormer, chen2022dual_DamFormer, cui2023mtscd_MTSCD} introduce hierarchical transformer networks, which are both based on the swin transformer \citep{liu2021swin}. The others typically follow a paradigm in which features extracted by CNN serve as semantic tokens, followed by contextual relation modeling between bi-temporal tokens using transformer.

In contrast to the above, our proposed method considers the change detection and captioning tasks as a video understanding process. It integrates learnable perception frames to efficiently capture changes utilizing a video encoder, thus obviating the necessity for meticulously designed change extractors and providing a unified framework.

\textbf{Video Understanding.}
Video understanding in action recognition \citep{huang2024froster_FROSTER, feichtenhofer2020x3d_X3D, feichtenhofer2019slowfast_SlowFast} and detection \citep{li2022uniformer_UniFormerV1, liu2022video_VideoSwin, fan2021multiscale_MViTv1, li2022mvitv2_MViTv2} involves the analysis of video data to identify and localize actions performed by objects within the video frames. Over the past years, video understanding advances have mostly been driven by 3D convolutional neural networks (CNNs) \citep{tran2015learning_C3D, feichtenhofer2019slowfast_SlowFast, ji20123d_3DCNN}. However, they suffer from difficult optimization and large computation cost problems. To resolve the issue, I3D \citep{carreira2017quo_I3D} inflates the pre-trained 2D convolution kernels for better optimization. Methods in \citep{qiu2017learning_P3D, tran2018closer_R2Plus1D, tran2019video_CSN, feichtenhofer2020x3d_X3D, wang2020video_CorrNet} try to factorize 3D convolution kernel in different dimensions to reduce complexity, others \citep{lin2019tsm_TSM, wang2016temporal_TSN, jiang2019stm_STM, li2020tea_TEA} introduce well-designed temporal modeling modules to enhance the temporal modeling ability for 2D CNNs, meanwhile saving the computation cost. Recently, several works \citep{TimeSFormer, fan2021multiscale_MViTv1, li2022mvitv2_MViTv2, liu2022video_VideoSwin, li2022uniformer_UniFormerV1, huang2024froster_FROSTER} have explored the potential of Vision Transformer \citep{dosovitskiy2020image_ViT} for spatiotemporal learning, verifying the outstanding ability to capture long-term dependencies. 

Building upon previous studies, we reconceptualize bi-temporal images as comprising two frames akin to a tiny video. Through the integration of perception frames, employing a video encoder facilitates efficient capture of changes.

\begin{figure*}[tb]
\centering
\includegraphics[width=\linewidth]{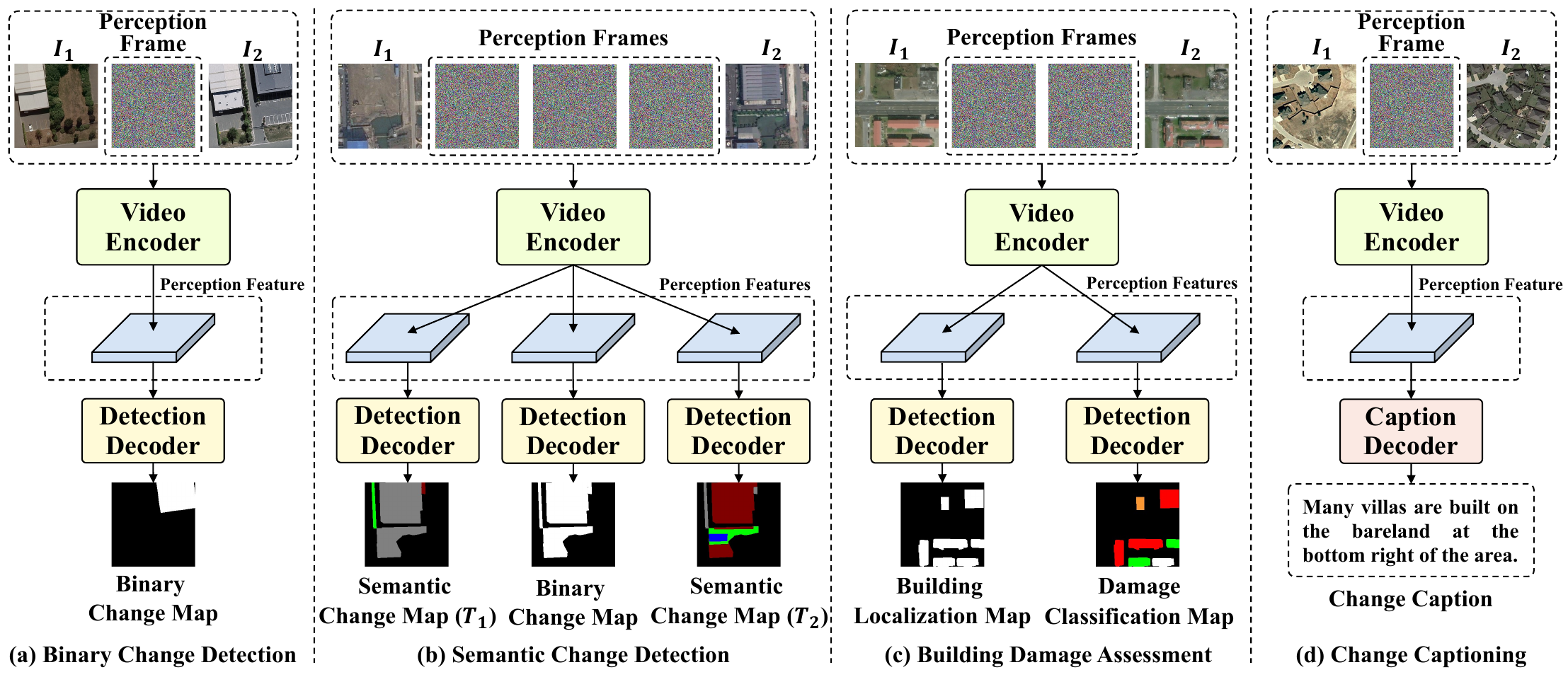}
\caption{\textbf{Overall architectures of Change3D for Binary Change Detection, Semantic Change Detection, Building Damage Assessment, and Change Captioning.} (a) Binary change detection necessitates acquiring a feature to represent changed targets, thus a perception frame is incorporated for sensing. (b) Semantic change detection involves representing semantic changes in $T_1$ and $T_2$ alongside binary changes. To accomplish this, three perception frames are integrated to facilitate semantic change learning. (c) Building damage assessment entails expressing two perception features for building localization and damage classification. Therefore, two perception frames are inserted to capture building damage. (d) Change captioning involves generating a feature that represents the altered content, thus incorporating a perception frame for interpreting content changes.
}
\label{fig:framework}
\end{figure*}

\section{Method}
\label{sec:method}
The overall architectures of Change3D are illustrated in Fig.~\ref{fig:framework}, comprising two main components: (1) a video encoder for extracting perception features, and (2) one or more decoders tasked with transforming perception features into change maps or captions. In the following, we first introduce the problem formulation in Sec.~\ref{subsec:problem_formulation}, followed by a detailed exposition of our method in Sec.~\ref{subsec:change3d}.

\subsection{Problem Formulation}
\label{subsec:problem_formulation}

In this paper, we focus on two key tasks within the remote sensing field: change detection, which includes binary change detection (BCD), semantic change detection (SCD), and building damage assessment (BDA); and change captioning (CC). These tasks are formulated as follows:

\textbf{Change Detection.}
Change detection involves identifying and analyzing changes between bi-temporal images captured at different times. Given two images $I_{1}, I_{2} \in \mathcal{R}^{H \times W \times C}$, the change detection process can be formulated as: $(\mathcal{M}_{\text{binary}}, \mathcal{M}_{\text{semantic}}, \mathcal{M}_{\text{damage}}) = \mathcal{F}_{\text{CD}}(I_{1}, I_{2})$. Here, $\mathcal{F}_{\text{CD}}$ is a change detector, $\mathcal{M}_{\text{binary}} \in \{0,1\}^{H \times W}$ represents the binary change map indicating change or no change, $\mathcal{M}_{\text{semantic}} \in \{0,1,\cdots,N_{\text{semantic}}-1\}^{H \times W}$ is the semantic change map classifying land cover into $N_{\text{semantic}}$ categories, and $\mathcal{M}_{\text{damage}} \in \{0,1,\cdots,N_{\text{damage}}-1\}^{H \times W}$ denotes the building damage map with $N_{\text{damage}}$ levels of damage.

\textbf{Change Captioning.}
Change captioning involves generating natural language descriptions that explain differences between bi-temporal images. Given a pair of bi-temporal images $I_{1}, I_{2} \in \mathcal{R}^{H \times W \times C}$, and a vocabulary set $\mathcal{V}$, change captioning process is expressed as: $\mathcal{W} = \mathcal{F}_{\text{CC}}(I_{1}, I_{2}) \in \mathcal{V}^L$. Here, $\mathcal{F}_{\text{CC}}(\cdot)$ is a change captioner, and $\mathcal{W}$ represents the sequence of words describing the changes, with $L$ being the length of the caption.

\subsection{Change3D}
\label{subsec:change3d}
To explore the versatility of our proposed method, we apply Change3D to change detection and captioning tasks. As depicted in Fig.~\ref{fig:framework}, we first initialize a set number of perception frames corresponding to the number of decoders for each task. Subsequently, these frames are stacked with bi-temporal images along temporal dimensions to compose the input video frames. Then, a video encoder is employed to facilitate the interaction between the perception frames and bi-temporal images to produce perception features. Finally, the corresponding decoders convert the perception features into change maps or captions.

\subsubsection{Perception Feature Extraction}
\label{subsubsec:perception_feature_extraction}
Given a pair of bi-temporal images $I_{1}$, $I_{2}$, along with learnable perception frames $I_{P}=\{I_{P}^{1},\cdots,I_{P}^{K}\}$, each with dimensions $(H, W, 3)$, where $K$ denotes the length of perception frames, with possible values of 1, 3, 2, or 1, the values corresponding to BCD, SCD, BDA, and CC, respectively. The perception frame functions similarly to the CLS token in Vision Transformers \cite{dosovitskiy2020image_ViT}, capturing essential information across the whole frame sequence. These frames are concatenated along the temporal dimension to construct a video. Then, a video encoder is employed to extract multi-layer features $f$, represented as \cref{eq:encoder}:
\begin{equation}
    \centering
    f = \mathcal{F}_{\text{enc}}(I_{1} \textcircled{c} I_{P} \textcircled{c} I_{2}),
    \label{eq:encoder}
\end{equation}
where $\mathcal{F}_{\text{enc}}(\cdot)$ is a video encoder, \textcircled{c} denotes concatenation along temporal dimension, $f \in \{\mathcal{R}^{T \times C_{i} \times \frac{H}{2^{i+1}} \times \frac{W}{2^{i+1}}}\}_{i=0}^{3}$, where $T=K+2$ and $C_{i}$ is the channel dimension. $f_{i}^{t}$ indicates the extracted feature of the $t$-th frame from the $i$-th encoder layer.
We use the multi-layer perception features to detect changes with diverse sizes, denoted as $p_{\text{det}}^{i,j} \in \{f_{i}^{1},\cdots,f_{i}^{T-2}\}$, where $p_{\text{det}}^{i,j}$ denotes the $j$-th perception feature in the $i$-th layer. And we leverage high-level perception feature $f_{3}^{1}$  to caption changes, denoted as $p_{\text{cap}}$.

To augment the representation capacity of perception features, a convolutional layer with the kernel size of 1$\times$1 followed by a ReLU is utilized to integrate the differential features of bi-temporal features $f_{i}^{0}$ and $f_{i}^{T-1}$ into them.

\subsubsection{Decoder and Optimization}
\label{subsubsec:ecoder_and_optimization}
\textbf{Change Decoder.} Existing methods \citep{feng2023change_DMINet, liu2023attention_AMTNet, peng2021scdnet_SCDNet, cui2023mtscd_MTSCD, chang2024triple_JFRNet, chen2024changemamba_ChangeMamba} utilize complex approaches for capturing bi-temporal changes and predicting change maps. To more convincingly highlight the learning prowess of Change3D, we choose to implement a simpler decoder. Specifically, we employ a cascade convolutional layer followed by an upsampling operation to progressively aggregate perception features from deep to shallow layers, ultimately restoring them to the original resolution of $H \times W$, as represented in \cref{eq:decoder}:
\begin{equation}
    \centering
    p_{\text{det}}^{i,j} \leftarrow \mathrm{Deconv_{4 \times 4}}(\mathrm{Conv_{1 \times 1}}(p_{\text{det}}^{i+1,j})) + p_{\text{det}}^{i,j},
    \label{eq:decoder}
\end{equation}
where $\mathrm{Conv_{1 \times 1}}$ is a 2D convolution with a kernel size of $1 \times 1$, $\mathrm{Deconv_{4 \times 4}}$ denotes transposed 2D convolution with a kernel size of $4 \times 4$ and stride of $2 \times 2$.

Finally, a classification layer is applied to transform the shallowest features $p_{\text{det}}^{0,j}$ into change maps $\mathcal{M}$, which is formulated as \cref{eq:cls_layer}:
\begin{equation}
    \centering
    \mathcal{M} = \mathop{\arg\max}(\mathrm{Conv_{3 \times 3}}(p_{\text{det}}^{0,j})),
    \label{eq:cls_layer}
\end{equation}
where $\mathrm{Conv_{3 \times 3}}$ is a 2D convolution with the kernel size of $3 \times 3$,  $\mathcal{M} \in \{0,\cdots,N-1\}^{H \times W}$, $N$ denotes the number of land cover categories or damage levels, and $\mathop{\arg\max}$ operation is performed along channel dimension.

\textbf{Caption Decoder.}
Following common practice, as adopted by \citep{liu2022remote_RSICCformer, liu2023decoupling_PromptCC, zhou2024single_SEN}, we employ a transformer-based \citep{dosovitskiy2020image_ViT} decoder to describe the changes. Specifically, the input words $w$ are initially passed through a word embedding layer along with a learnable positional embedding to obtain feature embeddings $e_w \in \mathcal{R}^{L \times D}$, where $L$ represents the word length and $D$ denotes the embedding dimension. Next, masked self-attention and cross-attention blocks are applied to model the changes based on the perception features, as described in \cref{eq:attn}:
\begin{equation}
    \centering
    e_{w}^{'} = \mathrm{CrossAttn}(\mathrm{SelfAttn}(e_w), p_{\text{cap}}) + e_w,
    \label{eq:attn}
\end{equation}
where the masking operation is performed on the attention matrix to prevent the leakage of words at subsequent positions. The deepest feature $p_{\text{cap}}$ serves as keys and values, while the feature embeddings $e_w$, enhanced by SelfAttn($\cdot$), function as queries.

Finally, a fully connected (FC) layer followed by a softmax function is applied to $e_{w}^{'}$ to predict the probability of change captions, which is represented as \cref{eq:fc}:
\begin{equation}
    \centering
    \mathcal{W} = \mathrm{softmax({FC}}(e_{w}^{'})).
    \label{eq:fc}
\end{equation}

Similar to existing methods \citep{zhang2023global_GASNet, li2024decoder_DEFO, zheng2021building_ChangeOS, liu2023decoupling_PromptCC}, we employ joint loss functions to optimize four tasks. Formulations of the loss functions are as follows:
\begin{align}
    \mathcal{L}_{\text{BCD}} &= \mathcal{L}_{\text{ce}} + \mathcal{L}_{\text{dice}}, ~
    \mathcal{L}_{\text{SCD}} = \mathcal{L}_{\text{ce}} + \mathcal{L}_{\text{dice}} + \mathcal{L}_{\text{sim}}, \nonumber \\
    \mathcal{L}_{\text{BDA}} &= \mathcal{L}_{\text{ce}} + \mathcal{L}_{\text{dice}}, ~
    \mathcal{L}_{\text{CC}} = \mathcal{L}_{\text{ce}},
\end{align}
where $\mathcal{L}_{\text{ce}}$, $\mathcal{L}_{\text{dice}}$, and $\mathcal{L}_{\text{sim}}$ denote cross-entropy, dice \citep{milletari2016v_dice}, and cosine similarity \citep{ding2022bi_Bi-SRNet} loss, respectively.


\section{Experiments}
\label{sec:experiments}
\subsection{Experimental Settings}
\label{subsec:experimental_settings}
\textbf{Datasets and Evaluation Metrics:}
We conduct extensive experiments on eight public datasets: LEVIR-CD \citep{chen2020spatial_levir_cd}, WHU-CD \citep{ji2018fully_whu_cd}, CLCD \citep{liu2022cnn_clcd}, HRSCD \citep{daudt2019multitask_HRSCD}, SECOND \citep{yang2021asymmetric_ASN_SECOND}, xBD \citep{gupta2019creating_xBD}, LEVIR-CC \citep{liu2022remote_RSICCformer}, and DUBAI-CC \citep{hoxha2022change_DUBAI}. The LEVIR-CD, WHU-CD, and CLCD are three binary change detection datasets, including changes in buildings, roads, and lakes. The HRSCD and SECOND are collected for semantic change detection, including five and six land-cover categories, respectively, for semantic learning. The xBD is a building damage assessment dataset that provides information on four damage categories. LEVIR-CC and DUBAI-CC are change captioning datasets that provide detailed descriptions of building and urbanization changes, with five sentences per image pair. We report change detection performance using F1 score, intersection over union (IoU), overall accuracy (OA), and SeK coefficient, and assess change captioning performance using BLEU-N (N=1,2,3,4), METEOR, ROUGE-L, and CIDEr.

\textbf{Implementation Details:}
We adopt I3D \citep{carreira2017quo_I3D}, Slow-R50 \citep{feichtenhofer2019slowfast_SlowFast}, UniFormer-XS \citep{li2022uniformer_UniFormerV1}, and X3D-L \citep{feichtenhofer2020x3d_X3D} as video encoders to verify the effectiveness of our method. Unless otherwise stated, we use X3D-L for our experiments, and we conduct diagnostic studies on the largest publicly available open-source dataset, xBD \citep{gupta2019creating_xBD}. All video encoders are pre-trained on Kinetics-400 \citep{kay2017kinetics_KINETICS}, with the decoder and perception frames initialized randomly. Additionally, we explore the initialization weights pre-trained on other video datasets in Tab.~\ref{tab:different_pretrained_models_on_bda}. Models are trained using the PyTorch \citep{paszke2019pytorch_PyTorch} framework on an NVIDIA GeForce RTX 3090 GPU. During training, we apply data augmentation through random resizing, flipping, and cropping. For optimization, we opt for the Adam optimizer \citep{kingma2014adam}, with beta values set to (0.9, 0.99) and a weight decay of 1e-4. Initially, the learning rate is 2e-4 and gradually reduces according to a scheduled reduction formula: (1-(curr\_iter/max\_iter))$^{\alpha}$ $\times$ lr, where $\alpha$ is set to 0.9, and max\_iter is set to 200K iterations for xBD, and 80K iterations for the rest datasets. The images of all datasets are cropped into 256$\times$256 non-overlap patches, with a batch size of 8 for HRSCD and SECOND, 32 for LEVIR-CC and DUBAI-CC, and 16 for the others.

\begin{table*}[htb]
\setlength\tabcolsep{2.5pt}
\centering
\caption{Performance comparison of different binary change detection methods on LEVIR-CD, WHU-CD, and CLCD datasets. $\dagger$ denotes the results with our reimplementation. The best results are \textbf{bolded} and the second-best results are \underline{underlined}. Values in parentheses indicate the percentage relative to the SOTA method (AMTNet). All results of the three evaluation metrics are described as percentages (\%).}
\scalebox{0.8}{
\begin{tabular}{c|c|c|c|ccc|ccc|ccc}
\hline
\multirow{2}{*}{Method} & \multirow{2}{*}{\#Params(M)} & \multirow{2}{*}{FLOPs(G)} & Inference & \multicolumn{3}{c|}{LEVIR-CD} & \multicolumn{3}{c|}{WHU-CD} & \multicolumn{3}{c}{CLCD} \\
\cline{5-13}
& & & (s/sample) & F1 & IoU & OA & F1 & IoU & OA & F1 & IoU & OA \\
\midrule
DTCDSCN \citep{liu2020building_DTCDSCN} & 41.07 (250\%) & 20.44 (83\%) & 0.029 & 87.43 & 77.67 & 98.75 & 79.92 & 66.56 & 98.05 & 57.47 & 40.81 & 94.59 \\
SNUNet \citep{fang2021snunet_SNUNet} & 12.04 (73\%) & 54.82 (222\%) & 0.036 & 88.16 & 78.83 & 98.82 & 83.22 & 71.26 & 98.44 & 60.82 & 43.63 & 94.90 \\
ChangeFormer \citep{bandara2022transformer_ChangeFormer} & 41.03 (250\%) & 202.79 (822\%) & 0.081 & 90.40 & 82.48 & 99.04 & 87.39 & 77.61 & 99.11 & 61.31 & 44.29 & 94.98 \\
BIT \citep{chen2021remote_BIT} & \underline{3.55} (22\%) & \underline{10.63} (43\%) & 0.027 & 89.31 & 80.68 & 98.92 & 83.98 & 72.39 & 98.52 & 59.93 & 42.12 & 94.77 \\
ICIFNet \citep{feng2022icif_ICIF} & 23.82 (145\%) & 25.36 (103\%) & 0.049 & 89.96 & 81.75 & 98.99 & 88.32 & 79.24 & 98.96 & 68.66 & 52.27 & 95.77 \\
Changer $\dagger$ \citep{fang2023changer_Changer} & 11.39 (69\%) & 11.86 (48\%) & \underline{0.025} & 90.70 & 82.99 & \underline{99.08} & 89.18 & 80.48 & 99.17 & 67.07 & 50.46 & 95.69 \\
DMINet \citep{feng2023change_DMINet} & 6.24 (38\%) & 14.42 (58\%) & 0.036 & 90.71 & 82.99 & 99.07 & 88.69 & 79.68 & 98.97 & 67.24 & 50.65 & 95.21 \\
GASNet $\dagger$ \citep{zhang2023global_GASNet} & 23.59 (143\%) & 23.52 (95\%) & 0.028 & 90.52 & 83.48 & 99.07 & 91.75 & 84.76 & \underline{99.34} & 63.84 & 46.89 & 94.01 \\
AMTNet \citep{liu2023attention_AMTNet} & 16.44 (100\%) & 24.67 (100\%) & 0.032 & 90.76 & 83.08 & 98.96 & \underline{92.27} & \underline{85.64} & 99.32 & \underline{75.10} & \underline{60.12} & \underline{96.45} \\
EATDer $\dagger$ \citep{ma2024eatder_EATDer} & 6.61 (40\%) & 23.43 (95\%) & 0.034 & \underline{91.20} & \underline{83.80} & 98.75 & 90.01 & 81.97 & 98.58 & 72.01 & 56.19 & 96.11 \\
\midrule
\textbf{Change3D} & \textbf{1.54} (9\%) & \textbf{8.29} (34\%) & \textbf{0.015} & \textbf{91.82} & \textbf{84.87} & \textbf{99.17} & \textbf{94.56} & \textbf{89.69} & \textbf{99.57} & \textbf{78.03} & \textbf{63.97} & \textbf{96.87} \\
\hline
\end{tabular}
}
\label{tab:overall_results_bcd}
\end{table*}

\begin{table*}[htb]
\setlength\tabcolsep{2.5pt}
\centering
\caption{Performance comparison of different semantic change detection methods on HRSCD and SECOND datasets. $\dagger$ denotes the results with our reimplementation. The best results are \textbf{bolded} and the second-best results are \underline{underlined}. Values in parentheses indicate the percentage relative to the SOTA method (Bi-SRNet). All results of the four evaluation metrics are described as percentages (\%).}
\scalebox{0.8}{
\begin{tabular}{c|c|c|c|cccc|cccc}
\hline
\multirow{2}{*}{Method} & \multirow{2}{*}{\#Params(M)} & \multirow{2}{*}{FLOPs(G)} & Inference & \multicolumn{4}{c|}{HRSCD} & \multicolumn{4}{c}{SECOND} \\
\cline{5-12}
& & & (s/sample) & F1 & mIoU & OA & SeK & F1 & mIoU & OA & SeK \\
\midrule
HRSCD-S4 \citep{daudt2019multitask_HRSCD} & 13.71 (59\%) & 43.69 (23\%) & 0.024 & 69.39 & 67.79 & 81.32 & 22.50 & 58.21 & 71.15 & 86.62 & 18.80 \\
ChangeMask $\dagger$ \citep{zheng2022changemask_ChangeMask} & \underline{8.73} (37\%) & \underline{37.16} (20\%) & \underline{0.021} & 70.59 & 67.56 & 81.65 & 23.43 & 59.74 & 71.46 & 86.93 & 19.50 \\
SCDNet \citep{peng2021scdnet_SCDNet} & 39.62 (169\%) & 116.98 (62\%) & 0.032 & 70.80 & 67.43 & 81.46 & 23.61 & 60.01 & 70.97 & \underline{87.40} & 19.73 \\
SSCD-L $\dagger$ \citep{ding2022bi_Bi-SRNet} & 23.39 (100\%) & 189.57 (100\%) & 0.029 & 69.69 & 66.21 & 81.05 & 21.88 & 61.22 & 72.60 & 87.19 & 21.86 \\
Bi-SRNet $\dagger$ \citep{ding2022bi_Bi-SRNet} & 23.39 (100\%) & 189.91 (100\%) & 0.031 & \underline{71.72} & \underline{67.83} & \underline{82.06} & \underline{24.59} & 61.85 & 72.08 & 87.20 & 21.36 \\
MTSCD \citep{cui2023mtscd_MTSCD} & 94.62 (405\%) & 290.28 (153\%) & 0.028 & 67.56 & 63.71 & 72.06 & 9.03 & 60.23 & 71.68 & 87.04 & 20.57\\
JFRNet \citep{chang2024triple_JFRNet} & 23.29 (100\%) & 47.38 (25\%) & 0.035 & 66.65 & 64.65 & 76.20 & 18.78 & \underline{62.63} & \underline{72.82} & 87.10 & \underline{22.56} \\
DEFO $\dagger$ \citep{li2024decoder_DEFO} & 26.02 (111\%) & 401.09 (211\%) & 0.025 & 66.49 & 65.67 & 81.36 & 19.20 & 61.18 & 72.39 & 87.01 & 21.08 \\
\midrule
\textbf{Change3D} & \textbf{1.66} (7\%) & \textbf{15.19} (8\%) & \textbf{0.018} & \textbf{73.29} & \textbf{68.67} & \textbf{82.57} & \textbf{26.85} & \textbf{62.83} & \textbf{72.95} & \textbf{87.42} & \textbf{22.98} \\
\hline
\end{tabular}
}
\label{tab:overall_results_scd}
\vspace{-0.8em}
\end{table*}

\subsection{Comparison with State-of-the-Art Methods}
\label{subsec:comparison_with_state-of-the-art_methods}
Tab.~\ref{tab:overall_results_bcd}-\ref{tab:overall_results_levir_cc_dubai_cc} represent comparisons to state-of-the-art methods on change detection and captioning tasks across eight datasets. Notably, all the compared methods employ a shared-weight image encoder with a change extractor for feature extraction. Specifically, BIT, ICIFNet, Changer, DMINet, GASNet, AMTNet, Bi-SRNet, DEFO, JFRNet, ChangeOS, PCDASNet, RSICCformer, and SEN apply ResNet \citep{he2016deep_ResNet} or its variants \citep{hu2018squeeze_senet} as image encoder, SNUNet and EATDer apply UNet \citep{ronneberger2015u_unet} and stacked non-local blocks \citep{wang2018non_nonlocal}, respectively. In contrast, our approach employs a video encoder to directly capture changes, eliminating the need for specialized change extractors and providing a unified framework for multiple change detection and captioning tasks. From Tab.~\ref{tab:overall_results_bcd}-\ref{tab:overall_results_levir_cc_dubai_cc}, we can summarize the following key findings: (1) Change3D achieves superior or competitive performance compared to the existing methods across all datasets and evaluation metrics, which demonstrates its effectiveness. (2) Compared with state-of-the-art methods, our proposed method requires $\sim$6\%-13\% fewer parameters, $\sim$8\%-34\% FLOPs, and has the fastest inference speed, while obtaining comparable performance over all datasets. For example, when compared with AMTNet, which introduces well-designed attention modules for bi-temporal feature interaction, our proposed method achieves significantly better results with little computation cost (9\%) and fewer parameters (34\%), indicating the efficiency of Change3D. (3) Our proposed method is adaptable to BCD, SCD, BDA, and CC without necessitating architectural redesign, while also achieving superior performance. This demonstrates the generalizability of Change3D across diverse change detection and captioning tasks.

\begin{table*}[tb]
\setlength\tabcolsep{2.5pt}
\centering
\caption{Performance comparison of different building damage assessment methods on xBD dataset. The best results are \textbf{bolded} and the second-best results are \underline{underlined}. Values in parentheses indicate the percentage relative to the SOTA method (PCDASNet). All results of the seven evaluation metrics are described as percentages (\%).}
\scalebox{0.8}{
\begin{tabular}{c|c|c|c|ccc|cccc}
\hline
\multirow{2}{*}{Method} & \multirow{2}{*}{\#Params(M)} & \multirow{2}{*}{FLOPs(G)} & Inference & \multirow{2}{*}{F$_{1}^{loc}$} & \multirow{2}{*}{F$_{1}^{cls}$} & \multirow{2}{*}{F$_{1}^{overall}$} & \multicolumn{4}{c}{Damage F$_{1}$ Per-class} \\
\cline{8-11}
& & & (s/sample) & & & & Non & Minor & Major & Destroy \\
\midrule
xBD baseline \citep{gupta2019creating_xBD} & 25.72 (99\%) & \underline{23.15} (24\%) & 0.021 & 80.47 & 3.42 & 26.54 & 66.31 & 14.35 & 0.94 & 46.57 \\
Weber et al. \citep{weber2020building_MaskRCNN} & 48.29 (186\%) & 37.48 (39\%) & 0.029 & 83.60 & 70.02 & 74.10 & \underline{90.60} & 49.30 & 72.20 & 83.70 \\
ChangeOS-R18 \citep{zheng2021building_ChangeOS} & \underline{15.99} (62\%) & 69.65 (73\%) & \underline{0.019} & 84.62 & 69.87 & 74.30 & 88.61 & 52.10 & 70.36 & 79.65 \\
ChangeOS-R34 \citep{zheng2021building_ChangeOS} & 26.10 (100\%) & 74.50 (78\%) & 0.021 & 85.16 & 70.28 & 74.74 & 88.63 & 52.38 & 71.16 & 80.08 \\
ChangeOS-R50 \citep{zheng2021building_ChangeOS} & 53.02 (204\%) & 101.35 (106\%) & 0.025 & 85.41 & 70.88 & 75.24 & 88.98 & 53.33 & 71.24 & 80.60 \\
ChangeOS-R101 \citep{zheng2021building_ChangeOS} & 72.01 (277\%) & 111.10 (116\%) & 0.031 & 85.69 & 71.14 & 75.50 & 89.11 & 53.11 & 72.44 & 80.79 \\
DamFormer \citep{chen2022dual_DamFormer} & 31.38 (121\%) & 220.68 (230\%) & 0.065 & \textbf{86.86} & 72.81 & 77.02 & 89.86 & \underline{56.78} & 72.56 & 80.51 \\
PCDASNet \citep{wang2024pcdasnet_PCDASNet} & 26.00 (100\%) & 95.9 (100\%) & 0.020 & 85.48 & \underline{73.83} & \underline{77.33} & 90.12 & 55.67 & \underline{75.74} & \underline{83.91} \\
\midrule
\textbf{Change3D} & \textbf{1.60} (6\%) & \textbf{11.74} (12\%) & \textbf{0.016} & \underline{85.74} & \textbf{76.71} & \textbf{79.42} & \textbf{95.08} & \textbf{58.70} & \textbf{76.50} & \textbf{86.76} \\
\hline
\end{tabular}
}
\label{tab:overall_results_bda}
\end{table*}

\begin{table*}[h]
\setlength\tabcolsep{1.0pt}
\centering
\caption{Performance comparison of different change captioning methods on LEVIR-CC and DUBAI-CC datasets. The best results are \textbf{bolded} and the second-best results are \underline{underlined}. Values in parentheses indicate the percentage relative to the SOTA method (SEN). Abbreviations B, M, R, and C refer to BLEU, METEOR, ROUGE, and CIDEr, respectively.}
\scalebox{0.8}{
\begin{tabular}{c|c|c|c|ccccccc|ccccccc}
\hline
\multirow{2}{*}{Method} & \multirow{2}{*}{\#Params(M)} & \multirow{2}{*}{FLOPs(G)} & Inference & \multicolumn{7}{c|}{LEVIR-CC} & \multicolumn{7}{c}{DUBAI-CC} \\
\cline{5-18}
& & & (s/sample) & B-1 & B-2 & B-3 & B-4 & M & R & C & B-1 & B-2 & B-3 & B-4 & M & R & C \\
\midrule
DUDA \citep{park2019robust_DUDA} & 80.31 (201\%) & 20.28 (84\%) & 0.014 & 81.44  & 72.22 & 64.24 & 57.79 & 37.15 & 71.04 & 124.32 & 58.82 & 43.59 & 33.63 & 25.39 & 22.05 & 48.34 & 62.78 \\
MCCFormer-S \citep{qiu2021describing_MCCFormer} & 162.55 (407\%) & 25.09 (104\%) & 0.015 & 79.90  & 70.87  & 62.80  & 56.31  & 36.17  & 69.46  & 120.46 & 52.97  & 37.02  & 27.62  & 22.57  & 18.64  & 43.29  & 53.81 \\
MCCFormer-D \citep{qiu2021describing_MCCFormer} & 162.55 (407\%) & 25.09 (104\%) & 0.016 & 80.42  & 70.87  & 62.86  & 56.38  & 36.37  & 69.32  & 120.44 & 64.65  & 50.45  & 39.36  & 29.48  & 25.09  & 51.27  & 66.51 \\
RSICCformer \citep{liu2022remote_RSICCformer} & 172.80 (433\%) & 27.10 (114\%) & 0.010 & 84.72  & 74.96  & 67.52  & 62.11  & 38.80  & 74.22  & 132.62 & 67.92  & 53.61  & 41.37  & 31.28  & 25.41  & 51.96 & 66.54 \\
PromptCC \citep{liu2023decoupling_PromptCC} & 408.58 (1024\%) & \underline{19.88} (84\%) & 0.013 & 83.66  & 75.73  & 69.10  & 63.54  & 38.82  & 73.72  & \underline{136.44} & \underline{70.03} & \underline{58.41} & \textbf{49.44} & \textbf{40.32} & \underline{26.48} & \underline{55.82} & \underline{85.44} \\
SEN \citep{zhou2024single_SEN} & \underline{39.90} (100\%) & 24.04 (100\%) & \underline{0.008} & \underline{85.10} & \underline{77.05} & \underline{70.01} & \underline{64.09} & \underline{39.59} & \underline{74.57} & 136.02 & 64.12 & 50.41 & 40.28 & 31.01 & 23.67 & 48.19 & 65.15 \\
\midrule
\textbf{Change3D} & \textbf{5.05} (13\%) & \textbf{2.39} (10\%) & \textbf{0.007} & \textbf{85.81} & \textbf{77.81} & \textbf{70.57} & \textbf{64.38} & \textbf{40.03} & \textbf{75.12} & \textbf{138.29} & \textbf{72.25} & \textbf{58.68} & \underline{47.13} & \underline{36.80}  & \textbf{27.06} & \textbf{56.04}  & \textbf{86.19} \\
\hline
\end{tabular}
}
\label{tab:overall_results_levir_cc_dubai_cc}
\vspace{-0.8em}
\end{table*}


\subsection{Diagnostic Study}
\label{subsec:diagnostic_study}
\textbf{Effectiveness with different architectures.}
To investigate the effectiveness of the proposed Change3D with different 3D architectures, we conduct experiments using both CNN-based (\ie, I3D \citep{carreira2017quo_I3D}, Slow-R50 \citep{feichtenhofer2019slowfast_SlowFast}, and X3D-L \citep{feichtenhofer2020x3d_X3D}) and Transformer-based (\ie, UniFormer-XS \citep{li2022uniformer_UniFormerV1}) models. Key observations from Tab.~\ref{tab:different_3D_architectures_bda} include: (1) When integrated with our proposed method, all video models exhibit satisfactory performance, indicating the efficacy of Change3D regardless of model architectures. (2) The X3D-L outperforms all other models, possibly due to its use of expansion operations, which perform network expansion across various dimensions (temporal duration, spatial resolution, width, and depth), thus enhancing its inter-frame modeling capacity.

\begin{table}[htb]
\setlength\tabcolsep{1pt}
\centering
\caption{Study the effectiveness of the proposed method with different 3D architectures on the xBD dataset.}
\scalebox{0.8}{
\begin{tabular}{c|ccc|cccc}
\hline
\multirow{2}{*}{Method} & \multirow{2}{*}{F$_{1}^{loc}$} & \multirow{2}{*}{F$_{1}^{cls}$} & \multirow{2}{*}{F$_{1}^{overall}$} & \multicolumn{4}{c}{Damage F$_{1}$ Per-class} \\
\cline{5-8}
& & & & Non & Minor & Major & Destroy \\
\midrule
I3D \citep{carreira2017quo_I3D} & 84.45 & 74.32 & 77.36 & 94.57 & 55.30 & 73.87 & 86.01 \\
Slow-R50 \citep{feichtenhofer2019slowfast_SlowFast} & 85.52 & 74.36 & 77.71 & 94.81 & 54.67 & 74.03 & \textbf{87.34} \\
UniFormer-XS \citep{li2022uniformer_UniFormerV1} & 85.56 & 76.11 & 78.94 & 95.04 & 57.75 & 75.41 & 87.27 \\
X3D-L \citep{feichtenhofer2020x3d_X3D} & \textbf{85.74} & \textbf{76.71} & \textbf{79.42} & \textbf{95.08} & \textbf{58.70} & \textbf{76.50} & 86.76 \\
\hline
\end{tabular}
}
\label{tab:different_3D_architectures_bda}
\vspace{-0.8em}
\end{table}

\textbf{Impact of pre-trained weights.}
As illustrated in Tab.~\ref{tab:different_pretrained_models_on_bda}, we explore the impact of pre-trained weights on Change3D by employing diverse initialization methods for initializing the video encoder (defaulting to Slow-R50). These methods encompass random initialization and several publicly accessible pre-training weights within the action recognition domain, \ie, AVA \citep{gu2018ava_AVA}, Charades \citep{sigurdsson2016hollywood_Charades}, Something-Something V2 (SSv2) \citep{goyal2017something_something_something_v2}, and Kinetics-400 (K400) \citep{kay2017kinetics_KINETICS}. Key observations from the results are as follows: (1) Without utilizing pre-trained weights, Change3D exhibits inferior performance compared to others, highlighting the effectiveness and necessity of pre-trained weights. (2) Despite being trained on action recognition data, all initialized weights demonstrate superior performance, indicating the ability of Change3D to bridge the gap between disparate domains and effectively transfer knowledge from action recognition to the field of change detection and captioning. (3) By utilizing pre-training weights from K400, Change3D achieves optimal results. This is mainly due to the K400 dataset providing a larger volume of data and more diverse scenarios compared to the other three datasets, which enhances the representational capability of the pre-trained model.

\begin{table}[tb]
\setlength\tabcolsep{1.5pt}
\centering
\caption{Investigation on the impact of different pre-trained weights on the xBD dataset.}
\scalebox{0.8}{
\begin{tabular}{c|ccc|cccc}
\hline
\multirow{2}{*}{Pre-trained} & \multirow{2}{*}{F$_{1}^{loc}$} & \multirow{2}{*}{F$_{1}^{cls}$} & \multirow{2}{*}{F$_{1}^{overall}$} & \multicolumn{4}{c}{Damage F$_{1}$ Per-class} \\
\cline{5-8}
 & & & & Non & Minor & Major & Destroy \\
\midrule
Random Init & 81.00 & 71.54 & 74.38 & 94.11 & 51.99 & 71.78 & 82.53 \\
AVA \citep{gu2018ava_AVA} & 85.19 & 75.41 & 77.94 & 94.61 & 55.94 & 73.47 & 86.24 \\
Charades \citep{sigurdsson2016hollywood_Charades} & 85.31 & 75.51 & 78.05 & 94.65 & 56.13 & 74.02 & \textbf{86.83} \\
SSv2 \citep{goyal2017something_something_something_v2} & 84.76 & 74.52 & 77.89 & 94.77 & 55.13 & 73.37 & 86.35 \\
K400 \citep{kay2017kinetics_KINETICS} & \textbf{85.87} & \textbf{75.56} & \textbf{78.67} & \textbf{94.79} & \textbf{58.02} & \textbf{75.03} & 86.30 \\
\hline
\end{tabular}
}
\label{tab:different_pretrained_models_on_bda}
\vspace{-0.8em}
\end{table}

\textbf{Impact of perception frame insertion position.}
As depicted in Tab.~\ref{tab:perception_frame_insertion_position_on_bda}, this study investigates the effects of the insertion position of perception frames. Particularly, the \textit{sandwiched} perception frame positioned centrally exhibits optimal performance. This optimal performance is primarily attributed to the ability of video encoding to allow perception frames to interact effectively with adjacent bi-temporal images, thereby detecting differences and accurately capturing regions of change.

\begin{figure*}[htb]
\centering
\includegraphics[width=0.90\linewidth]{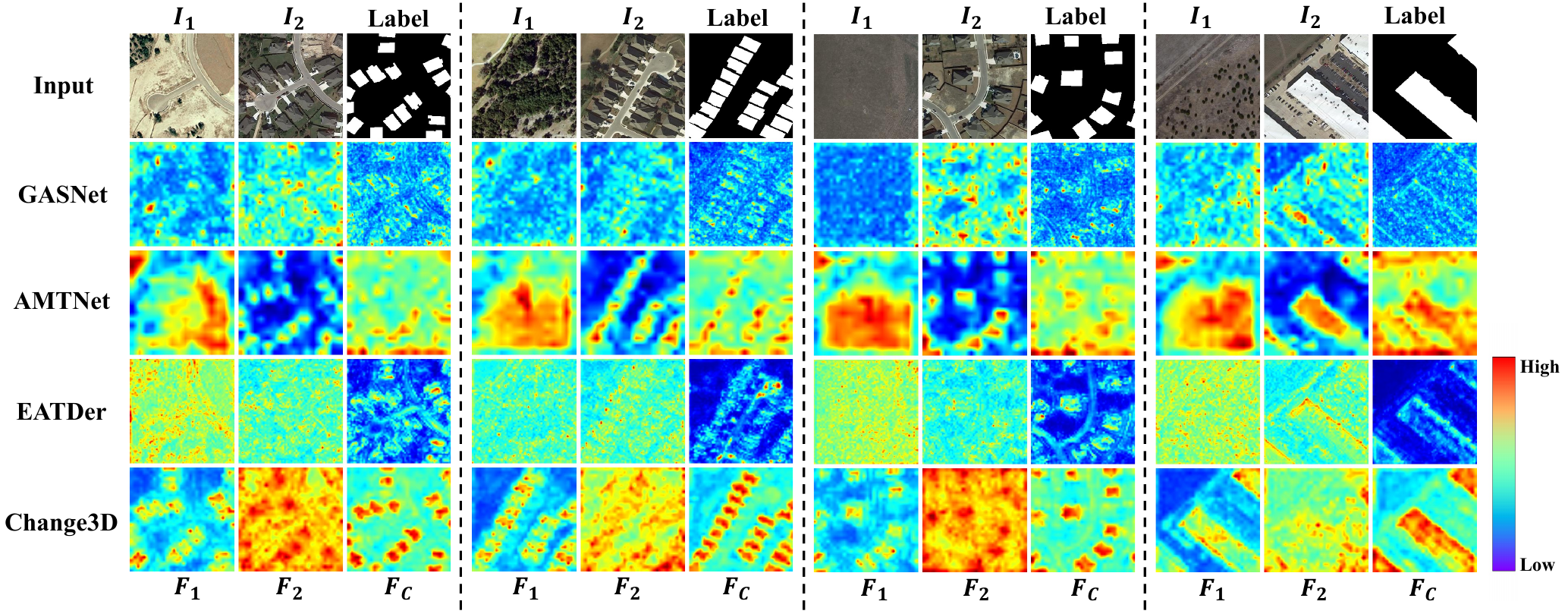}
\caption{
Visualization of bi-temporal features $F_1$, $F_2$, and extracted changes $F_C$. Our method directly focuses on changes during video encoding without intricate change extractors. The color bar on the right indicates the attention distribution for different colors.
}
\label{fig:attention_visualization}
\vspace{-0.8em}
\end{figure*}

\textbf{Impact of different ways obtaining perception features.}
We conduct four experiments to investigate the impact of various approaches for acquiring perception features. Specifically, FC represents a depth-wise 3D convolution with a kernel size of 3$\times$1$\times$1 and stride of 3$\times$1$\times$1 to aggregate the features. Mean and Max operations refer to average and max pooling along the temporal dimension, respectively. Feature selection involves choosing the perception feature based on the inserted index of the perception frame. Tab.~\ref{tab:various_acquisitions_for_perception_features_bda} illustrates that the selection methodology yields superior outcomes, effectively capturing changes through the interaction of features between perception frames and bi-temporal images.

\begin{table}[tb]
\setlength\tabcolsep{1.5pt}
\centering
\caption{Investigation on the impact of inserting perception frames $I_{P}^{1}$ and $I_{P}^{2}$ into different positions within the xBD dataset.}
\scalebox{0.75}{
\begin{tabular}{c|ccc|cccc}
\hline
\multirow{2}{*}{Position} & \multirow{2}{*}{F$_{1}^{loc}$} & \multirow{2}{*}{F$_{1}^{cls}$} & \multirow{2}{*}{F$_{1}^{overall}$} & \multicolumn{4}{c}{Damage F$_{1}$ Per-class} \\
\cline{5-8}
 & & & & Non & Minor & Major & Destroy \\
\midrule
$\{I_{P}^{1},I_{P}^{2},I_{1},I_{2}\}$ & 85.36 & 75.67 & 78.58 & 94.87 & 57.58 & 74.65 & 85.67 \\
$\{I_{1},I_{2},I_{P}^{1},I_{P}^{2}\}$ & 85.73 & 75.45 & 78.53 & 95.02 & 57.53 & 74.51 & 85.57 \\
$\{I_{1},I_{P}^{1},I_{P}^{2},I_{2}\}$ & \textbf{85.74} & \textbf{76.71} & \textbf{79.42} & \textbf{95.08} & \textbf{58.70} & \textbf{76.50} & \textbf{86.76} \\
\hline
\end{tabular}
}
\label{tab:perception_frame_insertion_position_on_bda}
\end{table}

\begin{table}[tb]
\setlength\tabcolsep{2.5pt}
\centering
\caption{Investigation on the impact of different ways of obtaining perception features on xBD dataset.}
\scalebox{0.75}{
\begin{tabular}{c|ccc|cccc}
\hline
\multirow{2}{*}{Method} & \multirow{2}{*}{F$_{1}^{loc}$} & \multirow{2}{*}{F$_{1}^{cls}$} & \multirow{2}{*}{F$_{1}^{overall}$} & \multicolumn{4}{c}{Damage F$_{1}$ Per-class} \\
\cline{5-8}
& & & & Non & Minor & Major & Destroy \\
\midrule
FC & 85.60 & 75.75 & 78.79 & 94.97 & 57.60 & 74.86 & 86.27 \\
Max & 85.57 & 75.30 & 78.47 & 94.94 & 56.63 & 74.55 & 86.52 \\
Mean & 85.64 & 74.56 & 77.88 & 94.79 & 55.42 & 74.68 & 85.68 \\
Selected & \textbf{85.74} & \textbf{76.71} & \textbf{79.42} & \textbf{95.08} & \textbf{58.70} & \textbf{76.50} & \textbf{86.76} \\
\hline
\end{tabular}
}
\label{tab:various_acquisitions_for_perception_features_bda}
\vspace{-0.8em}
\end{table}

\textbf{Effectiveness of perception frames.} 
We conduct experiments to investigate the effectiveness of integrating perception frames on HRSCD and SECOND datasets. The first line denotes that without incorporating perception frames, the output from the feature extractor is reshaped into feature maps that correspond to the number of perception features needed for the decoders. The second line represents the inclusion of perception frames. As shown in Tab.~\ref{tab:reshape_channel_to_frame}, performance improves with the integration of perception frames, demonstrating their effectiveness.

\textbf{Attention visualization.}
Fig.~\ref{fig:attention_visualization} illustrates attention maps from the last layer during image encoding and change extraction or video encoding, using four randomly selected samples from the LEVIR-CD dataset. Two significant observations emerge from the attention maps: (1) The perception feature of Change3D can efficiently discern changes through interaction with bi-temporal features, eliminating the need for dedicated change extractors. Conversely, other methods inadequately model change information during image feature encoding, underscoring the need for change extractors. (2) Moreover, for the extracted features from bi-temporal images, Change3D distinguishes itself from other methods by effectively concentrating on the changed regions. This capability ensures that the feature extraction process aligns with change detection and captioning tasks.

Please refer to \textit{Supplementary Material} for more experimental details, diagnostic studies, detailed architecture, theoretical analysis of video encoding, and qualitative results.


\begin{table}[tb]
\setlength\tabcolsep{1.5pt}
\centering
\caption{Study the effectiveness of perception frames on two semantic change detection datasets.}
\resizebox{1.0\linewidth}{!}{
\begin{tabular}{c|cccc|cccc}
\hline
\multirow{2}{*}{Method} & \multicolumn{4}{c|}{HRSCD} & \multicolumn{4}{c}{SECOND} \\
\cline{2-9}
& F$_{scd}$ & mIoU & OA & SeK & F$_{scd}$ & mIoU & OA & SeK \\
\midrule
Channel to Frames & 69.48 & 66.05 & 81.34 & 21.61 & 59.91 & 71.62 & 86.89 & 19.92 \\
\textit{w}/ Perception Frames & \textbf{73.29} & \textbf{68.67} & \textbf{82.57} & \textbf{26.85} & \textbf{62.83} & \textbf{72.95} & \textbf{87.42} & \textbf{22.98} \\
\hline
\end{tabular}
}
\label{tab:reshape_channel_to_frame}
\vspace{-0.8em}
\end{table}

\section{Conclusion}
\label{sec:conclusion}
This paper introduces a new paradigm, namely Change3D, to reconsider change detection and captioning from a video modeling perspective. It treats bi-temporal images as composed of two frames, resembling a miniature video. Through the integration of learnable perception frames between bi-temporal images, a video encoder with inter-frame modeling capabilities enables direct interaction between the perception frames and the images, allowing them to discern differences. Furthermore, this approach eliminates the need for complex, task-specific change extractors, providing a unified framework. We demonstrate the effectiveness of Change3D across eight well-recognized datasets with various model architectures. We hope the insight of this paper could inspire further study in more related computer vision tasks, \eg, depth completion and image matting.

\textbf{Acknowledge.} This work was supported by the National Key Research and Development Program of China (Grant No. 2023YFB3906102), the Fundamental Research Fund Program of LIESMARS (4201-420100071), and the Fundamental Research Program (WDZC20245250203).

\maketitlesupplementary

\section{More Experimental Details}
\label{sec:more_experimental_details}

\subsection{Dataset Description}
\textbf{Binary Change Detection Datasets}: 
(1) The LEVIR-CD \citep{chen2020spatial_levir_cd} comprises 637 bitemporal image pairs sourced from Google Earth, each with a high resolution of 0.5 m/pixel. Spanning images captured from 2002 to 2018 in various locations, this dataset includes annotations for 31333 individual building changes. (2) The WHU-CD \citep{ji2018fully_whu_cd} dataset focuses on building change detection and contains high-resolution (0.2 m/pixel) bi-temporal aerial images, totaling 32507$\times$15354 pixels. It primarily encompasses areas affected by earthquakes and subsequent reconstruction, mainly involving building renovations. (3) The CLCD \citep{liu2022cnn_clcd} dataset consists of cropland change samples, including buildings, roads, lakes, \etc. The bi-temporal images in CLCD were collected by Gaofen-2 in Guangdong Province, China, in 2017 and 2019, respectively, with spatial resolutions ranging from 0.5 to 2 m. Following the standard procedure detailed in \citep{zhang2023global_GASNet, liu2023attention_AMTNet}, each image of the three datasets is segmented into 256$\times$256 patches. Consequently, the LEVIR-CD dataset is divided into 7120 pairs for training, 1024 pairs for validation, and 2048 pairs for testing. The WHU-CD dataset is partitioned into 5947 training pairs, 744 validation pairs, and 744 test pairs. The CLCD dataset is divided into 1440, 480, and 480 pairs for training, validation, and testing, respectively.

\textbf{Semantic Change Detection Datasets}:
(1) The HRSCD \citep{daudt2019multitask_HRSCD} dataset contains a total of 291 image pairs of 10000$\times$10000 pixels, each with a resolution of 0.5 m/pixel. The images cover a range of urban and countryside areas in Rennes and Caen, France, including five classes of land cover, \ie, artificial surface, agricultural areas, forest, wetland, and water. (2) The SECOND \citep{yang2021asymmetric_ASN_SECOND} dataset consists of 4662 pairs of aerial images collected from several platforms and sensors, comprising of six land-cover categories, \ie, non-vegetated ground surface, tree, low-vegetation, water, buildings, and playgrounds, which are frequently involved in natural and man-made geographical changes. These pairs of images are distributed over various cities, including Hangzhou, Chengdu, and Shanghai. Considering that most of the labeled areas don't change in HRSCD, \eg, artificial surfaces and agricultural lands only account for 0.6\% of this dataset \citep{zhao2022spatially_SSESN}, we discard image pairs with less than 10\% of the pixels changed. Each image of the two datasets is cropped into 256$\times$256 non-overlap patches. Consequently, the HRSCD dataset is split into 6525, 932, and 1865 pairs for training, validation, and testing, and the SECOND dataset is divided into 11872 training pairs and 6776 testing pairs, respectively.

\textbf{Building Damage Assessment Dataset}:
The xBD \citep{gupta2019creating_xBD} is a large-scale building damage assessment dataset that provides high-resolution (0.8 m/pixel) satellite imagery with building localization and damage level labels, collected from 19 disaster events such as floods and earthquakes with an image size of 1024$\times$1024 pixels. The dataset uses polygons to represent building instances and provides four damage categories, \ie, non-damage, minor damage, major damage, and destroyed for each building. The minor damage pixels represent visible roof cracks or partially burnt structures while the major damage represents a partial wall, roof collapse, or structure surrounded by water. The destroyed label means that the building structure has completely collapsed, scorched or is no longer present. All the images are cropped into 256$\times$256 non-overlap patches, yielding 44785, 14928, and 14928 image pairs for training, holdout, and testing, respectively.

\textbf{Change Captioning Datasets}:
(1) The LEVIR-CC \citep{liu2022remote_RSICCformer} dataset is derived primarily from the LEVIR-CD \citep{chen2020spatial_levir_cd}, with each image having a spatial resolution of 1024$\times$1024 pixels and a resolution of 0.5 m/pixel. These bi-temporal images are sourced from 20 regions in Texas, USA, with a time span of 5 to 15 years. Each image pair is annotated with five sentences provided by five distinct annotators to describe the differences between the images. (2) The DUBAI-CC \citep{hoxha2022change_DUBAI} dataset focuses on urbanization changes in Dubai between 2000 and 2010. It contains 500 image tiles, each 50$\times$50 pixels, to analyze urbanization, extracted from bi-temporal images in the visible and infrared bands. It identifies six broad categories of change: roads, houses, buildings, green areas, lakes, and islands. Each bi-temporal image is annotated by five annotators, with each description containing at least three words addressing the spatial distribution and attributes of the changes. All images are cropped or resized into 256$\times$256 patches. The LEVIR-CC dataset is divided into 6815, 1333, and 1929 pairs for training, validation, and testing, respectively, and the DUBAI-CC dataset is split into 300 training pairs, 50 validation pairs, and 150 testing pairs.

\begin{table*}[tb]
\begin{minipage}{0.5\linewidth}
\setlength\tabcolsep{1.5pt}
\centering
\caption{Study the effectiveness of the proposed method with different 3D architectures on three binary change detection datasets, respectively.}
\resizebox{1.0\linewidth}{!}{
\begin{tabular}{c|ccc|ccc|ccc}
\hline
\multirow{2}{*}{Method} & \multicolumn{3}{c|}{LEVIR-CD} & \multicolumn{3}{c|}{WHU-CD} & \multicolumn{3}{c}{CLCD} \\
\cline{2-10}
& F1 & IoU & OA & F1 & IoU & OA & F1 & IoU & OA \\
\midrule
I3D \citep{carreira2017quo_I3D} & 91.21 & 83.84 & 99.11 & 94.18 & 89.01 & 99.55 & 78.67 & 64.84 & 96.92 \\
Slow-R50 \citep{feichtenhofer2019slowfast_SlowFast} & 91.39 & 84.14 & 99.13 & 94.37 & 89.34 & 99.56 & \textbf{78.82} & \textbf{65.05} & \textbf{96.93} \\
UniFormer-XS \citep{li2022uniformer_UniFormerV1} & 91.76 & 84.77 & 99.16 & 94.23 & 89.08 & 99.55 & 78.10 & 64.07 & 96.89 \\
X3D-L \citep{feichtenhofer2020x3d_X3D} & \textbf{91.82} & \textbf{84.87} & \textbf{99.17} & \textbf{94.56} & \textbf{89.69} & \textbf{99.57} & 78.03 & 63.97 & 96.87 \\
\hline
\end{tabular}
}
\label{tab:different_3D_architectures_bcd}
\end{minipage}
\hfill
\begin{minipage}{0.45\linewidth}
\setlength\tabcolsep{1.5pt}
\centering
\caption{Study the effectiveness of the proposed method with different 3D architectures on two semantic change detection datasets.}
\resizebox{1.0\linewidth}{!}{
\begin{tabular}{c|cccc|cccc}
\hline
\multirow{2}{*}{Method} & \multicolumn{4}{c|}{HRSCD} & \multicolumn{4}{c}{SECOND} \\
\cline{2-9}
& F$_{scd}$ & mIoU & OA & SeK & F$_{scd}$ & mIoU & OA & SeK \\
\midrule
I3D \citep{carreira2017quo_I3D} & 70.99 & 66.41 & 81.10 & 23.31  & 61.78 & 71.95 & 87.09 & 20.99 \\
Slow-R50 \citep{feichtenhofer2019slowfast_SlowFast} & 71.20 & 66.93 & 81.91 & 23.55 & 61.93 & 72.11 & 87.41 & 21.22 \\
UniFormer-XS \citep{li2022uniformer_UniFormerV1} & \textbf{73.69} & \textbf{69.30} & \textbf{83.07} & \textbf{27.31} & 62.00 & 71.79 & 87.03 & 21.22 \\
X3D-L \citep{feichtenhofer2020x3d_X3D} & 73.29 & 68.67 & 82.57 & 26.85 & \textbf{62.83} & \textbf{72.95} & \textbf{87.42} & \textbf{22.98} \\
\hline
\end{tabular}
}
\label{tab:different_3D_architectures_scd}
\end{minipage}
\end{table*}

\begin{table*}[tb]
\begin{minipage}{0.8\linewidth}
\setlength\tabcolsep{1.5pt}
\centering
\caption{Study the effectiveness of the proposed method with different 3D architectures on the LEVIR-CC and DUBAI-CC datasets. Abbreviations B, M, R, and C refer to BLEU, METEOR, ROUGE, and CIDEr, respectively.}
\scalebox{0.8}{
\begin{tabular}{c|ccccccc|ccccccc}
\hline
\multirow{2}{*}{Method} & \multicolumn{7}{c|}{LEVIR-CC} & \multicolumn{7}{c}{DUBAI-CC} \\
\cline{2-15}
& B-1 & B-2 & B-3 & B-4 & M & R & C & B-1 & B-2 & B-3 & B-4 & M & R & C \\
\midrule
I3D \citep{carreira2017quo_I3D} & 86.09 & 78.06 & 71.16 & 65.34 & 40.18 & 75.30 & 138.29 & \textbf{73.11} & \textbf{60.80} & \textbf{50.42} & \textbf{40.69} & \textbf{28.68} & \textbf{60.01} & \textbf{91.18} \\
Slow-R50 \citep{feichtenhofer2019slowfast_SlowFast} & 85.56 & 77.65 & 70.46 & 64.52 & 39.94 & 75.01 & 137.52 & 73.05 & 60.14 & 48.43 & 37.83 & 27.22 & 57.56 & 88.81 \\
UniFormer-XS \citep{li2022uniformer_UniFormerV1} & \textbf{86.75} & \textbf{78.84} & \textbf{71.68} & \textbf{65.58} & \textbf{40.86} & \textbf{75.98} & \textbf{140.15} & 70.27 & 57.11 & 45.70 & 35.26 & 25.96 & 54.03 & 81.66 \\
X3D-L \citep{feichtenhofer2020x3d_X3D} & 85.81 & 77.81 & 70.57 & 64.38 & 40.03 & 75.12 & 138.29 & 72.25 & 58.68 & 47.13 & 36.80 & 27.06 & 56.04 & 86.19 \\
\hline
\end{tabular}
}
\label{tab:different_3D_architectures_levir_cc_dubai_cc}
\end{minipage}
\hfill
\begin{minipage}{0.18\linewidth}
\centering
\captionof{table}{Quantitative evaluation of attention maps.}
\scalebox{0.8}{
\begin{tabular}{c|c}
\hline
Method & MSE \\
\midrule
GASNet \cite{zhang2023global_GASNet} & 0.35 \\
AMTNet \cite{liu2023attention_AMTNet} & 0.16 \\
EADTer \cite{ma2024eatder_EATDer} & 0.25 \\
\textbf{Change3D} & \textbf{0.07} \\
\hline
\end{tabular}
}
\label{tab:quantity_results_in_levir_cd}
\end{minipage}
\end{table*}

\subsection{Attention Visualization Details}
\label{subsec:attention_visualization_details_appendix}
As illustrated in Fig.~\ref{fig:attention_visualization}, to understand the feature distribution learned in each component within the models, we select three representative bi-temporal image-based methods for comparison, \ie, GASNet \citep{zhang2023global_GASNet}, AMTNet \citep{liu2023attention_AMTNet}, and EATDer \citep{ma2024eatder_EATDer}. The outputs $F_1$ and $F_2$ of these methods represent the final layer's output from the shared-weight image encoder at time $T_1$ and $T_2$, respectively. The differential features, represented as $F_C$, are extracted by the change extractor from the final layer to depict alterations. In our method, $F_1$ and $F_2$ correspond to the final layer's output during video feature encoding at time $T_1$ and $T_2$, respectively, while $F_C$ represents the perception features. For better visualization of the feature maps, we employ max and average pooling operations across the channel dimension to compress the features, then combine them via element-wise addition, and subsequently normalize them within the range of 0 to 1. 

In Tab.~\ref{tab:quantity_results_in_levir_cd}, we normalize the values in the attention map to [0, 1], apply a threshold of 0.5 to create binary maps, and calculate the MSE against the ground truth. Results show that our method achieves the lowest MSE and thus is more effective in focusing on changed regions.

\begin{table}[htb]
\setlength\tabcolsep{1.5pt}
\centering
\caption{Investigation on the impact of different initialization methods for perception frames across three binary change detection datasets.}
\resizebox{1.0\linewidth}{!}{
\begin{tabular}{c|ccc|ccc|ccc}
\hline
\multirow{2}{*}{Initialization} & \multicolumn{3}{c|}{LEVIR-CD} & \multicolumn{3}{c|}{WHU-CD} & \multicolumn{3}{c}{CLCD} \\
\cline{2-10}
& F1 & IoU & OA & F1 & IoU & OA & F1 & IoU & OA \\
\midrule
Zeros & 91.64 & 84.56 & 99.16 & 94.19 & 89.02 & 99.55 & 77.15 & 62.79 & 96.67 \\
Ones & 91.67 & 84.61 & 99.15 & 94.27 & 89.17 & 99.55 & 77.05 & 62.67 & 96.71 \\
Uniform & 91.75 & 84.77 & 99.16 & 94.43 & 89.45 & 99.56 & 77.76 & 63.61 & 96.80 \\
Random & \textbf{91.82} & \textbf{84.87} & \textbf{99.17} & \textbf{94.56} & \textbf{89.69} & \textbf{99.57} & \textbf{78.03} & \textbf{63.97} & \textbf{96.87} \\
\hline
\end{tabular}
}
\label{tab:different_initialization_on_perception_frames_bcd}
\end{table}

\begin{table}[htb]
\setlength\tabcolsep{1.5pt}
\centering
\caption{Investigation on the impact of different initialization methods for perception frames on the xBD dataset.}
\scalebox{0.8}{
\begin{tabular}{c|ccc|cccc}
\hline
\multirow{2}{*}{Initialization} & \multirow{2}{*}{F$_{1}^{loc}$} & \multirow{2}{*}{F$_{1}^{cls}$} & \multirow{2}{*}{F$_{1}^{overall}$} & \multicolumn{4}{c}{Damage F$_{1}$ Per-class} \\
\cline{5-8}
 & & & & Non & Minor & Major & Destroy \\
\hline
Zeros & 85.73 & 75.29 & 78.42 & 94.95 & 57.35 & 73.40 & 86.50 \\
Ones & 85.75 & 75.51 & 78.58 & 95.00 & 57.12 & 74.38 & 86.36 \\
Uniform & 85.96 & 75.67 & 78.76 & 94.94 & 57.05 & 75.00 & 86.43 \\
Random & \textbf{85.74} & \textbf{76.71} & \textbf{79.42} & \textbf{95.08} & \textbf{58.70} & \textbf{76.50} & \textbf{86.76} \\
\hline
\end{tabular}
}
\label{tab:perception_frame_initialization_on_bda}
\end{table}

\section{More Diagnostic Experiments}
\label{sec:diagnostic_study_appendix}

\textbf{Effectiveness with different architectures.}
The effectiveness of the proposed Change3D with different 3D architectures on BCD, SCD, and CC tasks is presented in \cref{tab:different_3D_architectures_bcd}-\ref{tab:different_3D_architectures_levir_cc_dubai_cc}. Notably, all video models exhibit competitive performance, underscoring the efficacy of the proposed method with various video modeling architectures.

\textbf{Impact of initialization on perception frames.}
We explore the effects of various initialization methods on perception frames by employing four different approaches: initializing with zeros, ones, uniform values between 0 and 1, and random initialization (\ie, with a mean of 0 and a standard deviation of 1). Analysis of Tab.~\ref{tab:different_initialization_on_perception_frames_bcd}-\ref{tab:perception_frame_initialization_on_bda} reveals that fixed-value initialization is less effective compared to the other methods. Random initialization yields the most favorable outcomes, which is reasonable as it enables a more robust generation of perception features.


\begin{table}[tb]
\setlength\tabcolsep{2.5pt}
\centering
\caption{Investigation on the impact of different similarity loss functions across two semantic change detection datasets.}
\scalebox{0.8}{
\begin{tabular}{c|cccc|cccc}
\hline
Similarity & \multicolumn{4}{c|}{HRSCD} & \multicolumn{4}{c}{SECOND} \\
\cline{2-9}
Loss & F$_{scd}$ & mIoU & OA & SeK & F$_{scd}$ & mIoU & OA & SeK \\
\midrule
- & 72.28 & 67.74 & 82.01 & 25.02 & 61.49 & 71.97 & 86.86 & 21.08 \\
L1 & 73.13 & 68.50 & 82.58 & 26.32 & 61.67 & 72.14 & 87.09 & 21.28 \\
L2 & 72.61 & 68.17 & 82.11 & 25.84 & 61.93 & 72.16 & 87.08 & 21.39 \\
Contrastive & 72.97 & 68.54 & 82.63 & 26.16 & 62.03 & 72.21 & 87.16 & 21.55 \\
Angular & 73.28 & \textbf{68.73} & 82.66 & 26.65 & \textbf{62.64} & 72.65 & 87.26 & 22.58 \\
Cosine & \textbf{73.29} & 68.59 & \textbf{82.74} & \textbf{26.73} & 62.61 & \textbf{72.84} & \textbf{87.40} & \textbf{22.86} \\
\hline
\end{tabular}
}
\label{tab:similarity_loss}
\end{table}

\textbf{Impact of different similarity losses.}
We conduct extensive ablation experiments to explore the impact of similarity losses, including L1, L2, Contrastive (with a margin of 0.5), Angular, and Cosine. Tab.~\ref{tab:similarity_loss} presents several key observations: (1) Without the similarity loss, Change3D exhibits inferior performance compared to others, highlighting the effectiveness and necessity of the similarity loss for semantic change detection task. (2) Cosine and Angular losses outperform others on both HRSCD and SECOND datasets, as they better handle changes in content rather than intensity or scale. L1, L2, and Contrastive losses are more sensitive to outliers, potentially skewing results, with Contrastive loss also being highly sensitive to the margin.

\begin{table}[tb]
\setlength\tabcolsep{1.5pt}
\centering
\caption{Performance comparison of several representative methods with random initialization across three binary change detection datasets.}
\resizebox{1.0\linewidth}{!}{
\begin{tabular}{c|ccc|ccc|ccc}
\hline
\multirow{2}{*}{Method} & \multicolumn{3}{c|}{LEVIR-CD} & \multicolumn{3}{c|}{WHU-CD} & \multicolumn{3}{c}{CLCD} \\
\cline{2-10}
& F1 & IoU & OA & F1 & IoU & OA & F1 & IoU & OA \\
\midrule
GASNet \citep{zhang2023global_GASNet} & 89.38 & 82.59 & 98.86 & 90.85 & 82.36 & 99.06 & 60.83 & 42.35 & 92.50 \\
AMTNet \citep{liu2023attention_AMTNet}  & 88.94 & 80.08 & 98.89 & 90.23 & 79.27 & 98.78 & 69.32 & 52.17 & 95.29 \\
EATDer \citep{ma2024eatder_EATDer} & 89.35 & 82.31 & 98.86 & 88.79 & 80.01 & 99.07 & 69.46 & 53.35 & 94.68 \\
\hline
\textbf{Change3D} & \textbf{90.80} & \textbf{83.16} & \textbf{99.08} & \textbf{92.40} & \textbf{85.88} & \textbf{99.41} & \textbf{71.55} & \textbf{55.71} & \textbf{96.11} \\
\hline
\end{tabular}
}
\label{tab:random_init_for_bcd_compared_methods}
\end{table}

\begin{table}[tb]
\setlength\tabcolsep{1.5pt}
\centering
\caption{Performance comparison of several representative methods with random initialization on the xBD dataset.}
\resizebox{1.0\linewidth}{!}{
\begin{tabular}{c|ccc|cccc}
\hline
\multirow{2}{*}{Method} & \multirow{2}{*}{F$_{1}^{loc}$} & \multirow{2}{*}{F$_{1}^{cls}$} & \multirow{2}{*}{F$_{1}^{overall}$} & \multicolumn{4}{c}{Damage F$_{1}$ Per-class} \\
\cline{5-8}
 & & & & Non & Minor & Major & Destroy \\
\midrule
ChangeOS-R101 \citep{zheng2021building_ChangeOS} & 80.31 & 67.74 & 72.11 & 88.80 & 46.86 & 67.30 & 75.09 \\
DamFormer \citep{chen2022dual_DamFormer} & 80.70 & 69.48 & 72.19 & 88.23 & 49.53 & 68.02 & 78.34 \\
PCDASNet \citep{wang2024pcdasnet_PCDASNet} & 80.05 & 68.79 & 72.36 & 90.05 & 48.76 & 71.69 & 78.84 \\
\hline
\textbf{Change3D} & \textbf{81.00} & \textbf{71.54} & \textbf{74.38} & \textbf{94.11} & \textbf{51.99} & \textbf{71.78} & \textbf{82.53} \\
\hline
\end{tabular}
}
\label{tab:random_init_for_bda_compared_methods}
\end{table}

\begin{figure}[tb]
    \centering
    \includegraphics[width=0.7\linewidth]{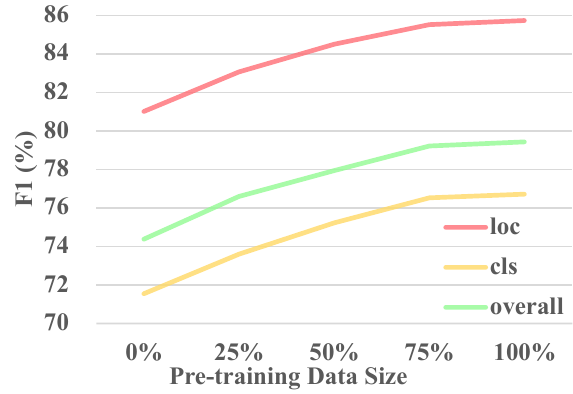}
    \caption{Pre-training data size \vs performance on the xBD dataset.}
    \label{fig:pre_training_data_ratio_vs_performance_on_bda}
\end{figure}

\textbf{Impact of pre-training \vs performance.}
(1) Since most 2D-model-based methods are typically initialized with ImageNet pre-training, our method is pre-trained using video data, such as K400 \citep{kay2017kinetics_KINETICS}, and SSv2 \citep{goyal2017something_something_something_v2}, \etc. To eliminate the influence of pre-training, we compare Change3D with several 2D-model-based methods using random initialization, as depicted in Tab.~\ref{tab:random_init_for_bcd_compared_methods}-\ref{tab:random_init_for_bda_compared_methods}. The table shows that \textbf{under the identical initialization setting}, Change3D consistently outperforms other approaches, highlighting the superiority of the proposed method.
(2) Most current state-of-the-art methods use pre-trained weights to initialize visual encoders, \eg, ImageNet1K (1.2M images) for AMTNet \cite{liu2023attention_AMTNet}, DEFO \cite{li2024decoder_DEFO}, and SEN \cite{zhou2024single_SEN}, and CLIP-400M (400M image-text pairs) for PromptCC \cite{liu2023decoupling_PromptCC}. Our video encoders use pre-trained datasets like K400 (1.9M images), AVA (1.6M images), and SSv2 (1.34M images), which are comparable to ImageNet1K but much less than CLIP-400M. (3) Fig.~\ref{fig:pre_training_data_ratio_vs_performance_on_bda} shows that performance improves with more pre-training data from K400 \cite{kay2017kinetics_KINETICS}, saturating after 75\%.
(4) Tab.~\ref{tab:different_pretrained_dataset_bcd} for the BCD task illustrates that pre-trained weights can improve model performance. Besides, pre-training on the K400 dataset still yields the best results, which is consistent with the findings from the BDA task of Tab.~\ref{tab:different_pretrained_models_on_bda}.

\begin{table}[tb]
\setlength\tabcolsep{1.0pt}
\centering
\caption{Investigation on the impact of different pre-trained weights across three binary change detection datasets.}
\resizebox{0.9\linewidth}{!}{
\begin{tabular}{c|ccc|ccc|ccc}
\hline
\multirow{2}{*}{Pre-trained} & \multicolumn{3}{c|}{LEVIR-CD} & \multicolumn{3}{c|}{WHU-CD} & \multicolumn{3}{c}{CLCD} \\
\cline{2-10}
& F1 & IoU & OA & F1 & IoU & OA & F1 & IoU & OA \\
\midrule
Random Init & 90.80 & 83.16 & 99.08 & 92.40 & 85.88 & 9.41 & 71.55 & 55.71 & 96.11 \\
AVA & 91.27 & 83.93 & 99.12 & 94.26 & 89.14 & 99.55 & 78.61 & 64.76 & 97.01 \\
Charades & 91.23 & 83.87 & 99.11 & 94.05 & 88.77 & 99.54 & 77.92 & 63.83 & 96.89 \\
SSv2 & 91.16 & 83.75 & 99.11 & 94.26 & 89.14 & 99.55 & 77.89 & 63.78 & 96.86 \\
K400 & \textbf{91.39} & \textbf{84.14} & \textbf{99.13} & \textbf{94.37} & \textbf{89.34} & \textbf{99.56} & \textbf{78.82} & \textbf{65.05} & \textbf{96.93} \\
\hline
\end{tabular}
}
\label{tab:different_pretrained_dataset_bcd}
\end{table}

\textbf{Necessity of multiple perception frames.}
Using multiple perception frames improves the model's capacity to learn individual characteristics for each sub-task. Results in Tab.~\ref{tab:single_vs_multi_perception_frame_on_scd} demonstrate that using multiple perception frames leads to superior results, highlighting their effectiveness.

\begin{table}[h]
\setlength\tabcolsep{1.0pt}
\centering
\caption{Single \vs multiple perception frames on two semantic change detection and one damage assessment datasets.}
\resizebox{1.0\linewidth}{!}{
\begin{tabular}{c|cccc|cccc|ccc}
\hline
Perception & \multicolumn{4}{c|}{HRSCD} & \multicolumn{4}{c|}{SECOND} & \multicolumn{3}{c}{xBD} \\
\cline{2-12}
Frame & F1 & mIoU & OA & SeK & F1 & mIoU & OA & SeK & F$_{1}^{loc}$ & F$_{1}^{cls}$ & F$_{1}^{overall}$ \\
\midrule
Single & 72.61 & 67.80 & 82.14 & 25.61 & 61.09 & 72.00 & 86.94 & 21.34 & 85.03 & 75.29 & 79.00 \\
Multiple & \textbf{73.29} & \textbf{68.67} & \textbf{82.57} & \textbf{26.85} & \textbf{62.83} & \textbf{72.95} & \textbf{87.42} & \textbf{22.98} & \textbf{85.74} & \textbf{76.71} & \textbf{79.42} \\
\hline
\end{tabular}
}
\label{tab:single_vs_multi_perception_frame_on_scd}
\end{table}

\begin{figure*}[tb]
\centering
\includegraphics[width=1.0\linewidth]{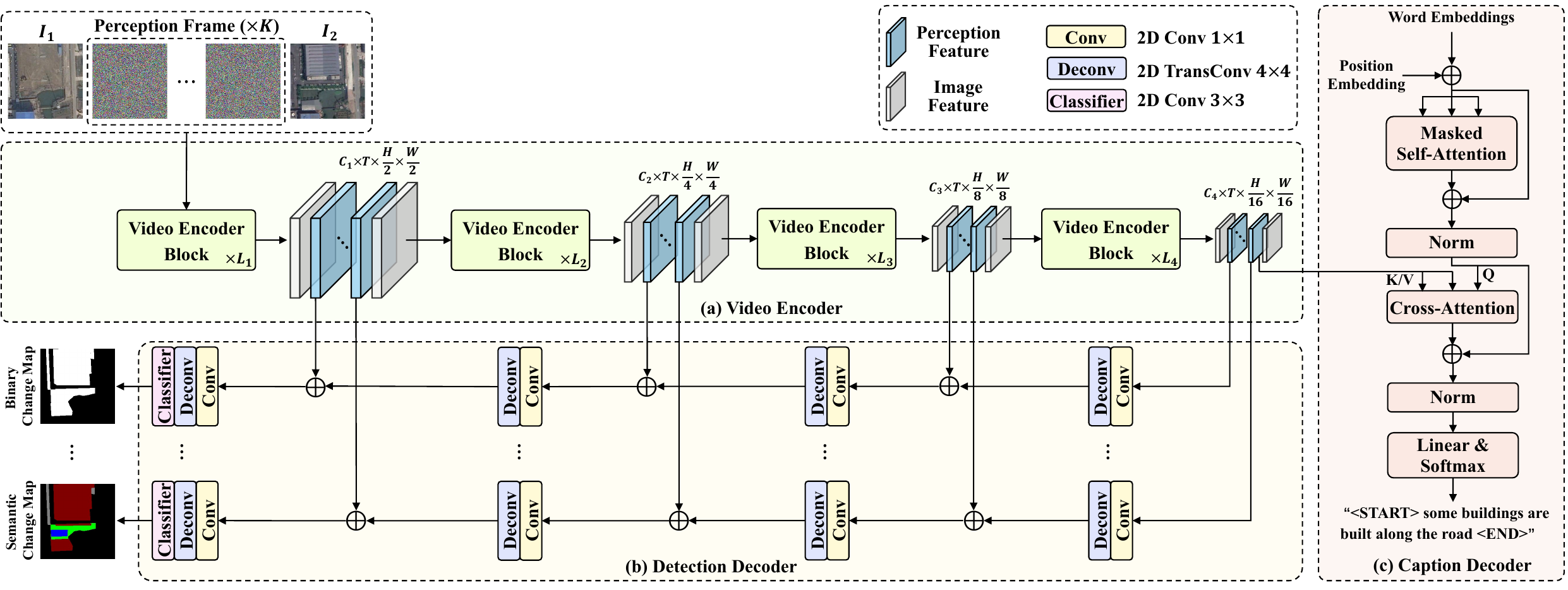}
\caption{Detailed architecture of Change3D. $L_{\{1-4\}}$ denote the number of blocks in each layer.
}
\label{fig:framework_v1.5_cd_cc_details}
\end{figure*}

\section{Detailed Architecture}
\label{sec:detailed_architecture}
As illustrated in Fig.~\ref{fig:framework_v1.5_cd_cc_details}, we provide a detailed architecture of Change3D, which is designed to address multiple tasks, including binary change detection, semantic change detection, building damage assessment, and change captioning. Each task involves a video encoder and task-specific decoders. Specifically, the input consists of bi-temporal images $I_1$ and $I_2$, along with $K$ perception frames that enhance temporal modeling by enriching inter-frame interactions. These inputs are stacked into a video frame sequence and processed through a multi-layer video encoder. The video encoder, equipped with spatiotemporal modeling capabilities, extracts robust features by integrating spatial details with temporal relationships, effectively capturing dynamic changes. The extracted features are then forwarded to task-specific decoders. For change map prediction, the framework leverages multi-layer perception features, while the highest-level semantic features from the encoder’s final layer are used for caption generation. Each decoder employs distinct parameters tailored to its respective task.

Change3D eliminates the need for complex, task-specific change extractors, providing a unified framework for diverse change detection and captioning tasks.

\section{Theoretical Analysis}
\label{sec:theoretical_analysis}
To establish a comprehensive theoretical foundation and illustrate the superiority of our proposed method, we present a theoretical analysis of video models applied to change detection and captioning tasks. Our approach diverges significantly from existing methods, particularly in differential feature extraction. Therefore, we provide a detailed analysis of the image encoding and change extraction in the previous paradigm, as well as the video encoding introduced in our proposed paradigm.

Our proposed video models (\ie, video encoder) can be conceptualized as a conditional probabilistic model that utilizes the video encoder’s inter-frame relation modeling capabilities to capture changes from the entire input frame sequence. Our method treats bi-temporal images and perception frames as sequences of video frames. Through video encoding, perception frames comprehensively capture contextual information among the images, thereby producing perception features for effective change representation.

\subsection{Previous Paradigm}
\label{subsec:previous_paradigm}
In the previous paradigm, bi-temporal image pairs are treated as separate inputs, each processed individually by a shared-weight image encoder to extract spatial features, followed by a dedicated change extractor. A decoder then makes predictions, as detailed below:
\begin{itemize}
    \item \textbf{Image Encoding:} Each image, $I_1$ and $I_2$, is independently encoded to produce feature representations $F_1$ and $F_2$:
    \begin{equation}
        \centering
        P(F_1 \mid I_1)~\text{and}~P(F_2 \mid I_2),
        \label{eq:image_encoding}
    \end{equation}
    where $P(F_1 \mid I_1)$ and $P(F_2 \mid I_2)$ describe the conditional probability distributions of extracting features $F_1$ and $F_2$ from images $I_1$ and $I_2$, respectively.

    \item \textbf{Change Extraction:} The differential features \( F_C \) are derived from a change extraction module:
    \begin{equation}
        \centering
        P(F_C \mid F_1, F_2),
        \label{eq:change_extraction}
    \end{equation}
    where this expression represents the conditional probability of obtaining the change features $F_C$, given the extracted features $F_1$ and $F_2$ from the two images.
    \item \textbf{Decoding:} The decoder transforms the differential features to change maps or captions $O$:
    \begin{equation}
        \centering
        P(O \mid F_C),
        \label{eq:decoding}
    \end{equation}
    where $P(O \mid F_C)$ denotes the conditional probability of generating output $O$, such as a change map or caption, based on the differential features $F_C$.
    \item \textbf{Joint Probability:} The joint probability for generating outputs given the inputs is influenced by independent image encoding and change extraction. Combined with \cref{eq:image_encoding,eq:change_extraction,eq:decoding}, we obtain:
    \begin{align}        
        P(O \mid I_1, I_2) &= P(F_1 \mid I_1) \cdot P(F_2 \mid I_2) \cdot P(F_C \mid F_1, F_2) \notag \\
        &\quad \cdot P(O \mid F_C),
        \label{eq:joint_probability}
    \end{align}
    which describes the process of generating the output conditioned on bi-temporal images, including the independent extraction of features and change detection.
    \item \textbf{Entropy:} The entropy expression is defined as follows:
    \begin{align}
        H(O, F_C, F_1, F_2)_{\text{prev}} &= H(O \mid F_C) + H(F_C \mid F_1, F_2) \notag \\
        &\quad + H(F_1 \mid I_1) + H(F_2 \mid I_2),
        \label{eq:conditional_entropy}
    \end{align}
    where each term represents the uncertainty at different stages: the entropy of the output given the change features, the entropy of the change features given the image features, and the entropy of each image feature conditioned on the respective input image.
    \item \textbf{Mutual Information:} The mutual information between the input and differential features is defined as:
    \begin{equation}
        \centering
        I(F_C; I_1, I_2)_{\text{prev}} = H(F_C)_{\text{prev}} + H(I_1, I_2) - H(F_C, I_1, I_2)
        \label{eq:mutual_information}
    \end{equation}
    which quantifies the information between the change features $F_C$ and the input images $I_1$ and $I_2$.
\end{itemize}

\subsection{Our Paradigm}
\label{subsec:our_paradigm}
Our approach redefines change detection and captioning tasks from a video modeling perspective. By incorporating learnable perception frames between the bi-temporal images, a video encoder facilitates direct interaction between the perception frame and the images to extract differences, as follows:
\begin{itemize}
    \item \textbf{Video Encoding:} The bi-temporal images $I_1$, $I_2$ incorporated with perception frames $I_P$ are stacked along the temporal dimension to construct a video, then a video encoder processes it to produce differential features $F_C$, which is formulated as follows:
    \begin{equation}
        \centering
        P(F_C | I_1, I_P, I_2),
        \label{eq:video_encoding}
    \end{equation}
    where this expression reflects the conditional probability of obtaining the differential features $F_C$ directly from the sequence of input images and the perception frame, effectively modeling the inter-frame relations for change extraction.
    \item \textbf{Decoding:} A decoder is applied to predict the change maps or captions $O$:
    \begin{equation}
        \centering
        P(O \mid F_C),
        \label{eq:video_decoding}
    \end{equation}
    where as before, describes the likelihood of generating output $O$ from the differential features $F_C$.
    \item \textbf{Joint Probability:} The joint probability benefits from holistic video encoding, which incorporates perception frames. Combined with \cref{eq:video_encoding,eq:video_decoding}, we get:
    \begin{equation}
        \centering
        P(O \mid I_1, I_P, I_2) = P(F_C | I_1, I_P, I_2) \cdot P(O \mid F_C),
        \label{eq:video_joint_probability}
    \end{equation}
    where this formulation emphasizes the integrated processing of temporal sequences through video encoding. This approach connects input frames with differential feature extraction seamlessly, eliminating the need for change extractor designs.
    \item \textbf{Entropy:} The entropy expression is formulated as:
    \begin{equation}
        H(O, F_C)_{\text{our}} = H(O \mid F_C) + H(F_C \mid I_1, I_P, I_2),
    \end{equation}
    which reflects the more efficient processing of information with reduced uncertainty across the change detection pipeline when using perception frames.
    \item \textbf{Mutual Information:} The mutual information between the input and differential features is defined as:
    \begin{align}
        I(F_C; I_1, I_2, I_P)_{\text{our}} &= H(F_C)_{\text{our}} + H(I_1, I_2, I_P) \notag \\
        &\quad - H(F_C, I_1, I_2, I_P),
    \end{align}
    illustrating enhanced mutual information and interdependence achieved by integrating perception frames within video encoding.
\end{itemize}

\subsection{Comparsion}
\label{subsec:comparsion}
\begin{itemize}
    \item \textbf{Probabilistic Model Comparison:} Our paradigm predicts output more accurately due to the inclusion of perception frames and holistic video encoding, which captures richer inter-frame information:
    \begin{equation}
        \centering
        P(O \mid I_1, I_2)_{\text{prev}} < P(O \mid I_1, I_P, I_2)_{\text{our}}
    \end{equation}
    \item \textbf{Entropy Comparison:} Our approach shows reduced overall entropy, indicating that the feature representations are more deterministic and less uncertain, leading to more reliable predictions.
    \begin{equation}
        \centering
        H(O, F_C, F_1, F_2)_{\text{prev}} > H(O, F_C)_{\text{our}}
    \end{equation}
    \item \textbf{Mutual Information Comparison:} Our paradigm captures a higher amount of information between the inputs and features, promoting enhanced understanding of changes as a result of direct interaction via video encoding.
    \begin{equation}
        \centering
        I(F_C; I_1, I_2)_{\text{prev}} < I(F_C; I_1, I_2, I_P)_{\text{our}}
    \end{equation}
\end{itemize}

\subsection{Summary}
\label{subsec:summary}
Our proposed paradigm demonstrates significant improvements in both probabilistic and information-theoretic measures. By incorporating perception frames into the video encoding process, our approach achieves:
\begin{itemize}
    \item Lower overall entropy, reflecting more deterministic feature representations.
    \item Higher mutual information, indicating better capture of the complex interdependencies and information among sequences.
    \item Enhanced joint probability model, delivering more accurate and reliable predictions in change detection and captioning tasks.
\end{itemize}

These advantages underscore the theoretical and empirical superiority of our paradigm, making it a simple yet effective framework for change detection and captioning tasks.

\section{Qualitative Results}
\label{sec:quality_results_appendix}
\begin{figure*}[tb]
\centering
\includegraphics[width=0.9\linewidth]{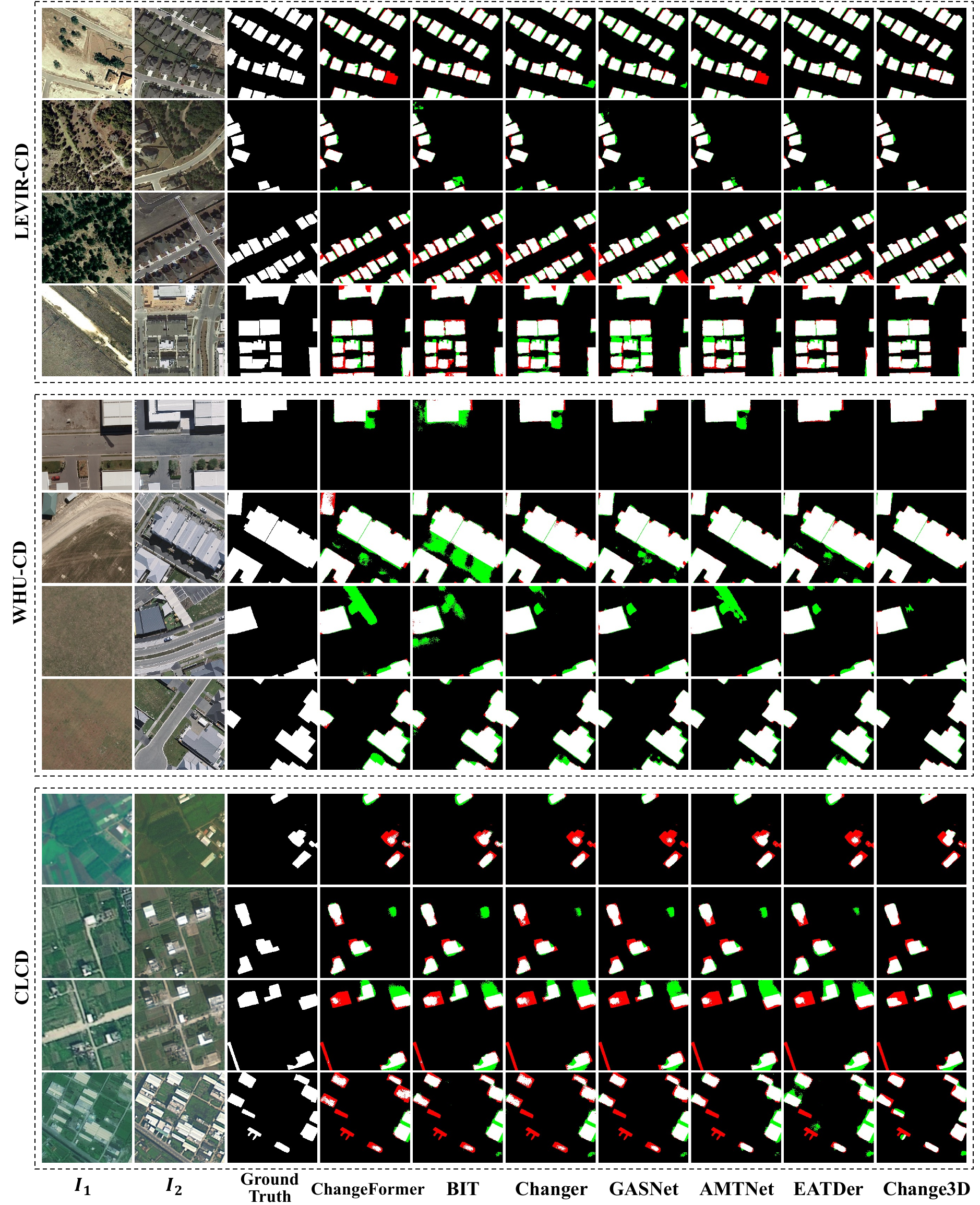}
\caption{Qualitative comparison of several representative methods on three binary change detection datasets, \ie, LEVIR-CD, WHU-CD, and CLCD. White represents a true positive, black is a true negative, \textcolor{green}{green} indicates a false positive, and \textcolor{red}{red} is a false negative. Fewer \textcolor{green}{green} and \textcolor{red}{red} pixels represent better performance.
}
\label{fig:pred_vis_bcd}
\end{figure*}

\begin{figure*}[htb]
\centering
\includegraphics[width=0.72\linewidth]{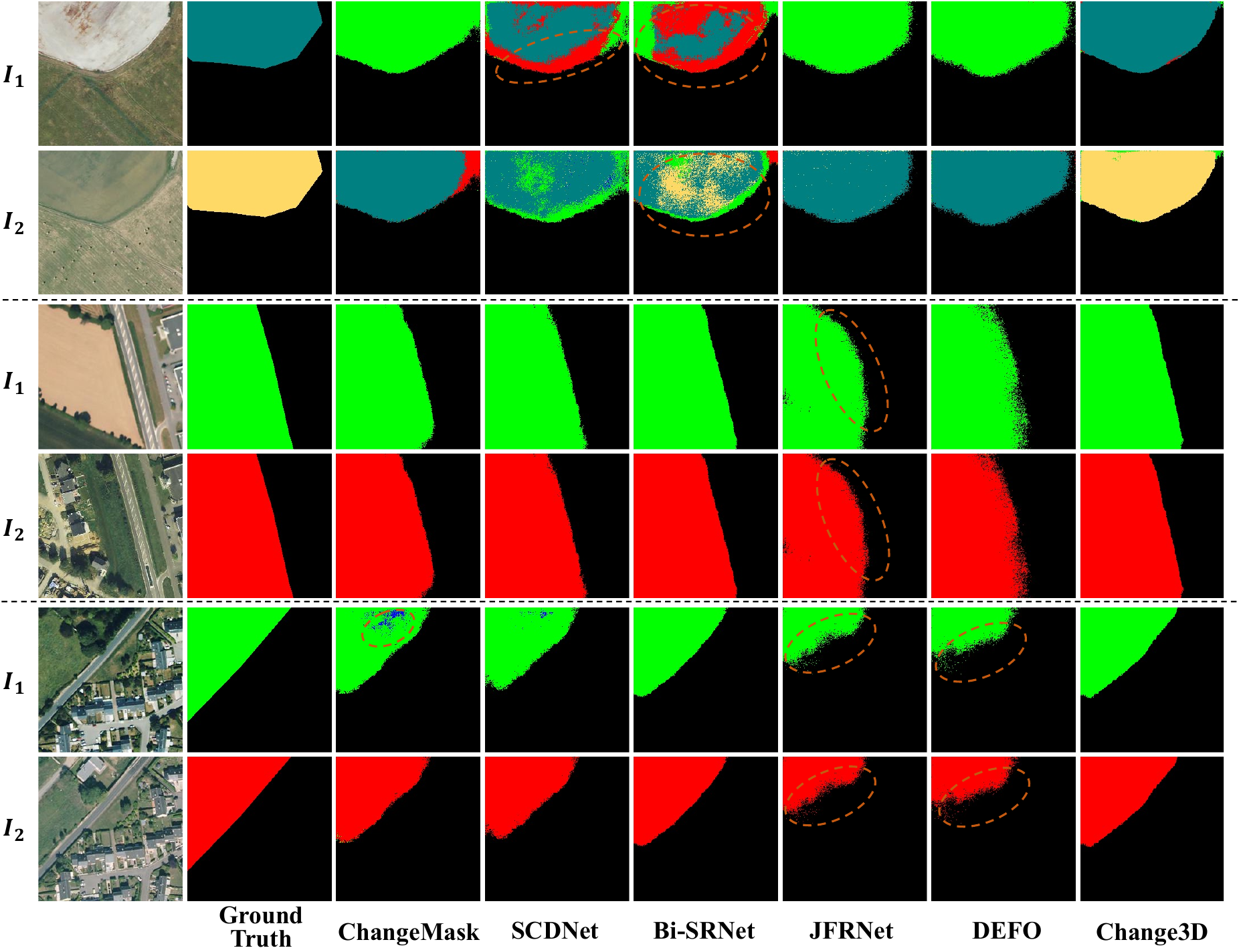}
\caption{Qualitative comparison of several representative methods on the HRSCD dataset. Black represents non-change, \textcolor{red}{red} denotes artificial surfaces, \textcolor{green}{green} indicates agricultural areas, \textcolor{blue}{blue} means forests, \textcolor{yellow}{yellow} represents wetlands, and \textcolor{teal}{teal} indicates water.
}
\label{fig:pred_vis_hrscd}
\end{figure*}

\begin{figure*}[htb]
\centering
\includegraphics[width=0.8\linewidth]{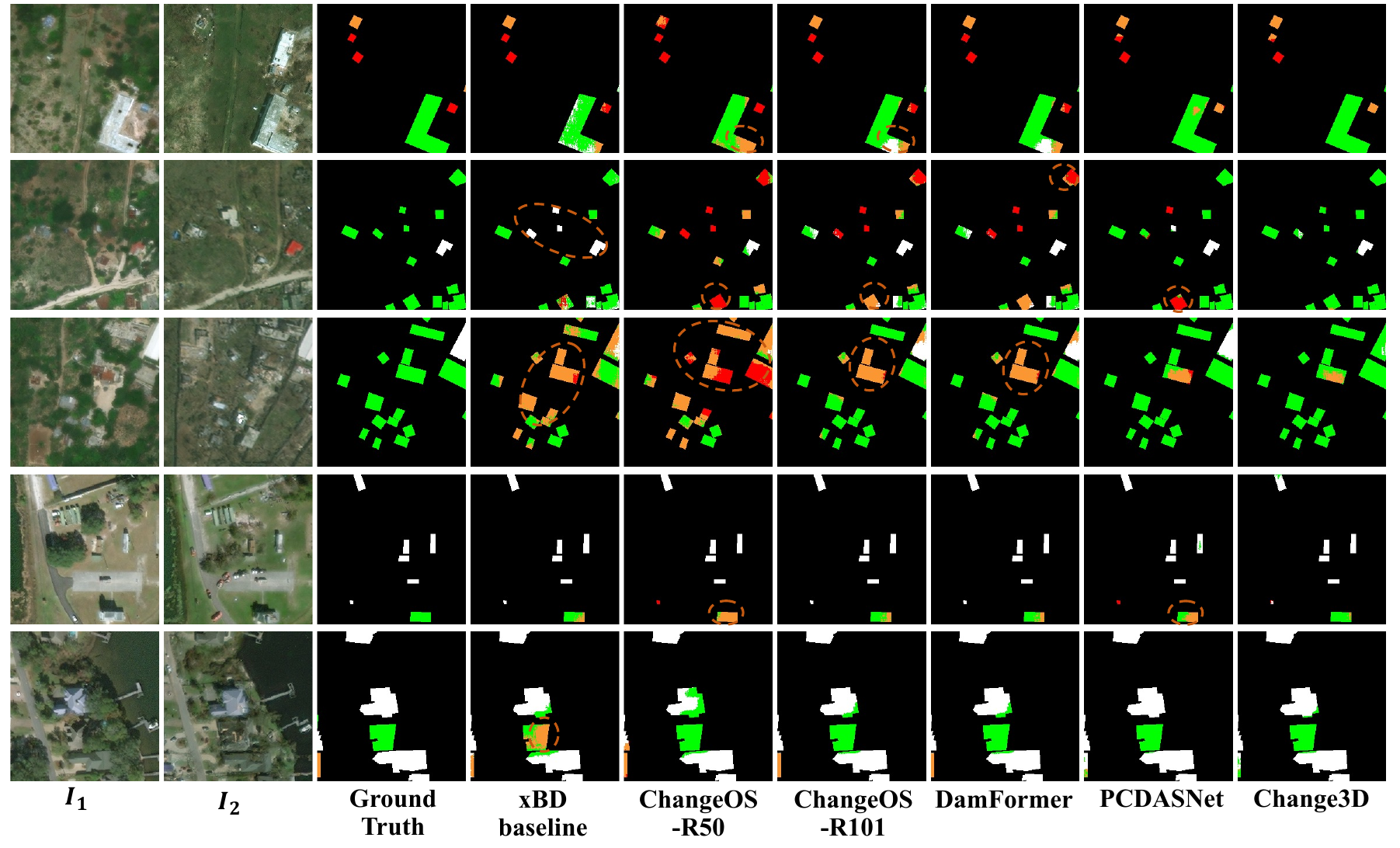}
\caption{Qualitative comparison of several representative methods on the xBD dataset. Black represents non-change, white denotes non-damage, \textcolor{green}{green} indicates minor damage, \textcolor{orange}{orange} represents major damage, \textcolor{red}{red} indicates destroyed.
}
\label{fig:pred_vis_xbd}
\end{figure*}

\begin{figure*}[htb]
\centering
\includegraphics[width=0.8\linewidth]{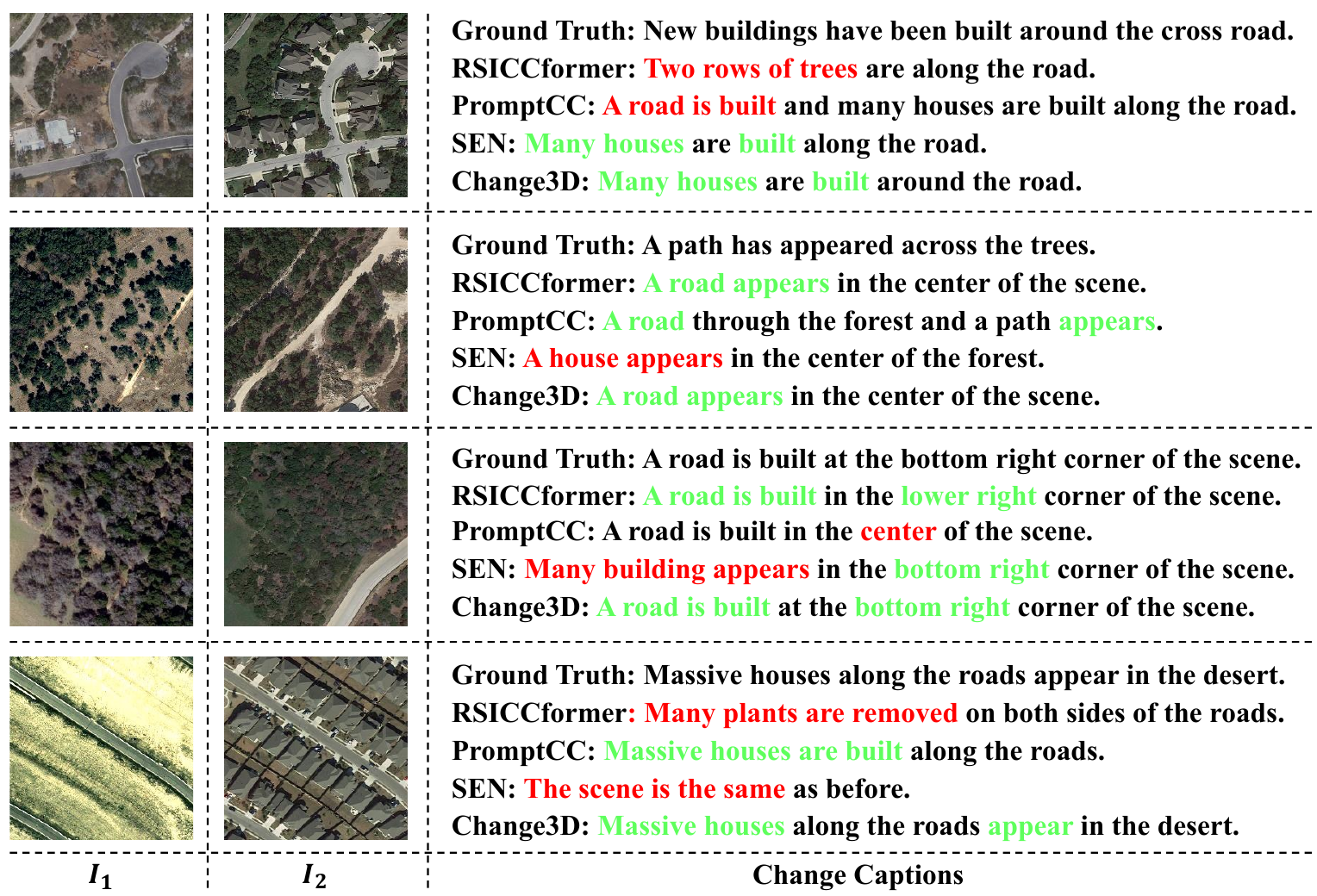}
\caption{Qualitative comparison of several representative methods on the LEVIR-CC dataset. \textcolor{green}{Green} indicates correct captions, while \textcolor{red}{red} indicates incorrect predictions.
}
\label{fig:pred_vis_levir_cc}
\end{figure*}

To qualitatively compare our method with previous approaches, we present comprehensive samples randomly selected from eight datasets, as illustrated in Fig.~\ref{fig:pred_vis_bcd}-\ref{fig:pred_vis_levir_cc}. Several key observations can be made from these samples: (1) Fig.~\ref{fig:pred_vis_bcd} shows that our proposed method outperforms all compared methods on the binary change detection task across various scenarios, including small, large, dense, and sparse changes. (2) Fig.~\ref{fig:pred_vis_hrscd}-\ref{fig:pred_vis_second} demonstrate that Change3D accurately identifies land cover types and produces clearer boundaries. (3) Fig.~\ref{fig:pred_vis_xbd} indicates that Change3D generates more accurate and semantically consistent assessment maps reflecting damage levels. (4) Fig.~\ref{fig:pred_vis_levir_cc} shows that Change3D provides more precise descriptions of the changes. These achievements are primarily attributed to the effective feature interaction between learnable perception frames and bi-temporal images in capturing differences, demonstrating the effectiveness of Change3D for multiple change detection and captioning tasks.

\definecolor{my_color_violet}{RGB}{255,153,255}

\begin{figure*}[htb]
\centering
\includegraphics[width=0.72\linewidth]{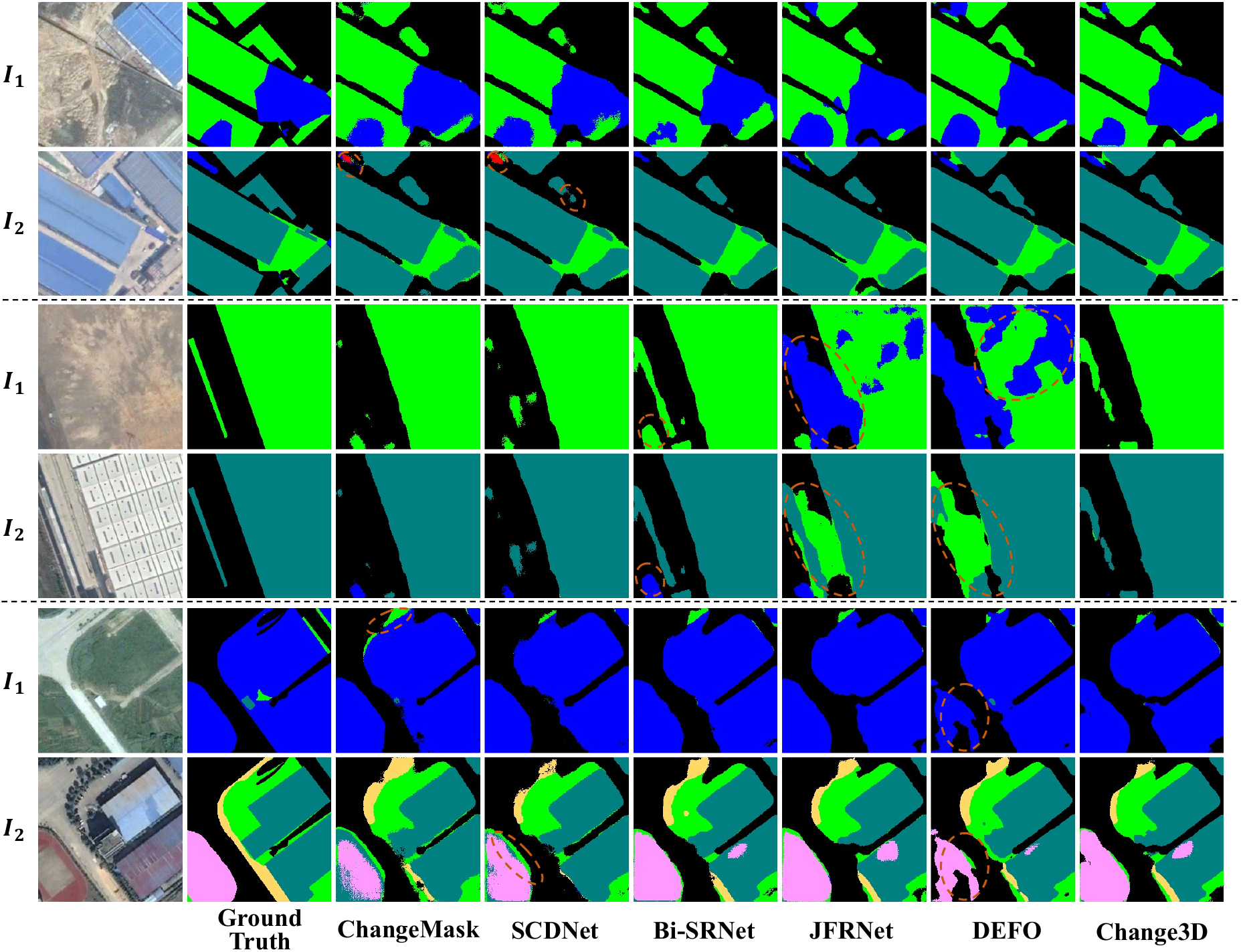}
\caption{Qualitative comparison of several representative methods on the SECOND dataset. Black represents non-change, \textcolor{red}{red} denotes low-vegetation, \textcolor{green}{green} indicates non-vegetated ground surface, \textcolor{blue}{blue} means trees, \textcolor{yellow}{yellow} represents water, \textcolor{teal}{teal} indicates buildings, and \textcolor{my_color_violet}{violet} denotes playgrounds.
}
\label{fig:pred_vis_second}
\end{figure*}

\clearpage

{
    \small
    \bibliographystyle{ieeenat_fullname}
    \bibliography{main}
}


\end{document}